\documentclass[sigconf]{acmart}

\usepackage{multirow}
\usepackage{color}
\usepackage{framed}
\usepackage{enumitem}

\AtBeginDocument{%
  }

\copyrightyear{2024}
\acmYear{2024}
\setcopyright{rightsretained}
\acmConference[KDD '24]{Proceedings of the 30th ACM SIGKDD Conference on Knowledge Discovery and Data Mining}{August 25--29, 2024}{Barcelona, Spain}
\acmBooktitle{Proceedings of the 30th ACM SIGKDD Conference on Knowledge Discovery and Data Mining (KDD '24), August 25--29, 2024, Barcelona, Spain}
\acmDOI{10.1145/3637528.3671576}
\acmISBN{979-8-4007-0490-1/24/08}

\makeatletter
\gdef\@copyrightpermission{
   \begin{minipage}{0.3\columnwidth}
     \href{https://creativecommons.org/licenses/by/4.0/}{\includegraphics[width=0.90\textwidth]{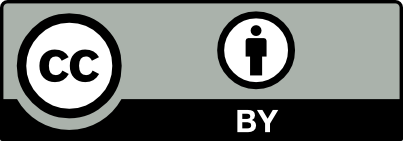}}
   \end{minipage}\hfill
   \begin{minipage}{0.7\columnwidth}
     \href{https://creativecommons.org/licenses/by/4.0/}{This work is licensed under a Creative Commons Attribution International 4.0 License.}
   \end{minipage}
   \vspace{5pt}
}
\makeatother

\begin{document}

\title{RareBench: Can LLMs Serve as Rare Diseases Specialists?}

\author{Xuanzhong Chen}
\authornote{Co-first authors.}
\orcid{0009-0002-1214-6130}
\affiliation{
    \department{Department of Computer Science and Technology \& Institute of Artificial Intelligence \& BNRist}
    \institution{Tsinghua University}
    \city{Beijing}
    \country{China}
}
\email{cxz23@mails.tsinghua.edu.cn}

\author{Xiaohao Mao}
\authornotemark[1]
\orcid{0009-0005-4636-4381}
\affiliation{
    \department{Department of Computer Science and Technology \& Institute of Artificial Intelligence \& BNRist}
    \institution{Tsinghua University}
    \city{Beijing}
    \country{China}
}
\email{mxh19@mails.tsinghua.edu.cn}

\author{Qihan Guo}
\authornotemark[1]
\orcid{0009-0003-6459-5525}
\affiliation{
    \department{Department of Computer Science and Technology \& Institute of Artificial Intelligence \& BNRist}
    \institution{Tsinghua University}
    \city{Beijing}
    \country{China}
}
\email{gqh22@mails.tsinghua.edu.cn}

\author{Lun Wang}
\orcid{0000-0001-5700-3622}
\affiliation{
    \department{Department of Internal Medicine}
    \institution{Peking Union Medical College Hospital}
    \institution{Chinese Academy of Medical Sciences \& Peking Union Medical College}
    \city{Beijing}
    \country{China}
}
\email{wanglun@pumch.cn}

\author{Shuyang Zhang}
\authornote{Corresponding authors.}
\orcid{0000-0002-1532-0029}
\affiliation{
    \department{Department of Cardiology}
    \institution{Peking Union Medical College Hospital}
    \institution{Chinese Academy of Medical Sciences \& Peking Union Medical College}
    \city{Beijing}
    \country{China}
}
\email{shuyangzhang103@nrdrs.org}

\author{Ting Chen}
\authornotemark[2]
\orcid{0000-0002-3228-9166}
\affiliation{
    \department{Department of Computer Science and Technology \& Institute of Artificial Intelligence \& BNRist}
    \institution{Tsinghua University}
    \city{Beijing}
    \country{China}
}
\email{tingchen@tsinghua.edu.cn}

\renewcommand{\shortauthors}{Xuanzhong Chen et al.}

\begin{abstract}
Generalist Large Language Models (LLMs), such as GPT-4, have shown considerable promise in various domains, including medical diagnosis. Rare diseases, affecting approximately 300 million people worldwide, often have unsatisfactory clinical diagnosis rates primarily due to a lack of experienced physicians and the complexity of differentiating among many rare diseases. In this context, recent news such as "ChatGPT correctly diagnosed a 4-year-old's rare disease after 17 doctors failed" underscore LLMs' potential, yet underexplored, role in clinically diagnosing rare diseases. To bridge this research gap, we introduce \textbf{\textit{RareBench}}, a pioneering benchmark designed to systematically evaluate the capabilities of LLMs on 4 critical dimensions within the realm of rare diseases. Meanwhile, we have compiled the largest open-source dataset on rare disease patients, establishing a benchmark for future studies in this domain. To facilitate differential diagnosis of rare diseases, we develop a dynamic few-shot prompt methodology, leveraging a comprehensive rare disease knowledge graph synthesized from multiple knowledge bases, significantly enhancing LLMs' diagnostic performance. Moreover, we present an exhaustive comparative study of GPT-4's diagnostic capabilities against those of specialist physicians. Our experimental findings underscore the promising potential of integrating LLMs into the clinical diagnostic process for rare diseases. This paves the way for exciting possibilities in future advancements in this field.
\end{abstract}



\maketitle

\section{Introduction}

Large Language Models (LLMs) like ChatGPT \cite{noauthor_introducing_nodate} have obtained widespread attention for their exceptional human-like language understanding and generation capabilities. 
As a result, applying LLMs in medicine is emerging as a promising research direction in artificial intelligence and clinical medicine. 
Several studies have been taken to explore how LLMs can assist doctors in various medical and clinical tasks, including medical diagnosis \cite{kwon2023large,wang2023chatcad}, clinical report generation \cite{yang2022large,singhal2023large}, and medicine education \cite{kung2023performance}. 
However, there is currently a lack of research investigating the capabilities and limitations of LLMs in the context of rare diseases.

Rare diseases collectively refer to a broad category of diseases, typically defined by their low prevalence in the population. Over 7,000 types of rare diseases are currently recognized \cite{haendel2020many}, with approximately 80\% being genetic in origin. 
Patients with rare diseases often face a high probability of misdiagnosis or underdiagnosis \cite{marwaha2022guide}, and the average time before receiving a confirmative diagnosis extends over several years \cite{evans2023dare}. 
The difficulty in diagnosis is largely attributed to the lack of prior exposure to rare diseases among physicians, hindering the accurate recognition of rare diseases and their associated phenotypes.  The phenotypes are typically symptoms, signs, or other disease-related information observed in rare disease patients that are used for disease diagnosis. However, there is significant phenotypic overlap among different rare diseases, as well as between rare diseases and common diseases, which further increases the difficulty of disease identification and diagnosis. 
Clinical diagnosis of rare diseases typically involves two primary steps. 
Initially, physicians collect clinical information from patients, including epidemiological information, symptoms, signs, past medical history, and family history, etc., to formulate an initial diagnosis. Next, specialized tests such as laboratory tests or imaging examinations will be conducted to further facilitate diagnosis and differential diagnosis. Additionally, due to the frequent involvement of numerous organs and systems in rare diseases, consulting specialists from different fields during the diagnostic process can help achieve a more comprehensive insight and final diagnosis. 

There have been many prior works to improve the diagnosis of rare diseases, including standardizing disease phenotype terminology into a hierarchical structure in the Human Phenotype Ontology (HPO) \cite{kohler2021human,kohler2017human,kohler2009clinical} and building knowledge bases annotating rare diseases with phenotypes such as the Online Mendelian Inheritance in Man (OMIM) \cite{hamosh2005online}, Orphanet \cite{ayme2003orphanet}, and the Compendium of China’s Rare Diseases (CCRD) \cite{he2019incidence}. 
These efforts result in a clear and structured representation of rare diseases: a disease or a patient can be represented by a set of associated phenotypes. From a machine learning perspective, computational methods can be developed to classify or rank diseases based on a patient’s phenotype information. 
These computational methods can be classified into two main categories. 
The first category treats the diagnosis of rare diseases as a ranking problem. 
In these approaches, a patient or a disease can be represented as phenotype vectors. Diseases are subsequently ranked by computing their semantic similarities with the patient. 
Methods falling into this category include PhenoMizer \cite{kohler2009clinical}, RDAD \cite{jia2018rdad}, RDD \cite{pinol2017rare}, and LIRICAL \cite{robinson2020interpretable}. 
However, these methods are constrained by their underlying assumptions, the lack of good-quality cases for training and testing, and the incomplete phenotypic information on many rare diseases in knowledge bases, often leading to relatively poorer diagnostic performance. 
The second category treats the diagnosis of rare diseases as an extreme multi-class classification task. 
Due to the scarcity of real-world data and the vast number of rare diseases to classify, this becomes a typical few-shot classification problem. 
Additionally, because of a lack of large-scale public rare disease patient datasets, most computational methods were only tested on simulated rare disease patient cases or small disease datasets with few diseases involved. Therefore, the clinical diagnostic capability of these methods remains unclear.

The prerequisite for diagnosing rare diseases using computational methods is to extract standardized and essential phenotypes/symptoms from electronic health records (EHRs) of clinical cases. 
To map clinical texts into standardized phenotypes, various natural language processing (NLP) methods have been developed, including EHR-Phenolyzer \cite{son2018deep}, ClinPhen \cite{deisseroth2019clinphen}, PhenoTagger \cite{luo2021phenotagger}, and PhenoBERT \cite{feng2022phenobert}. 
In a recent study, PhenoBCBERT and PhenoGPT \cite{yang2023enhancing} models were introduced to identify clinical phenotypes in clinical texts from pediatric patients. 
However, there are substantial variations in clinical texts in how physicians may record patient phenotypes, incorporating distinct details and terminology. 
Moreover, the current count of human phenotypes surpasses 17,000 \cite{kohler2021human}. 
All these factors present a significant challenge in precisely mapping or deducing phenotypes from clinical texts.

\begin{figure}[t]
  \includegraphics[width=0.8\columnwidth]{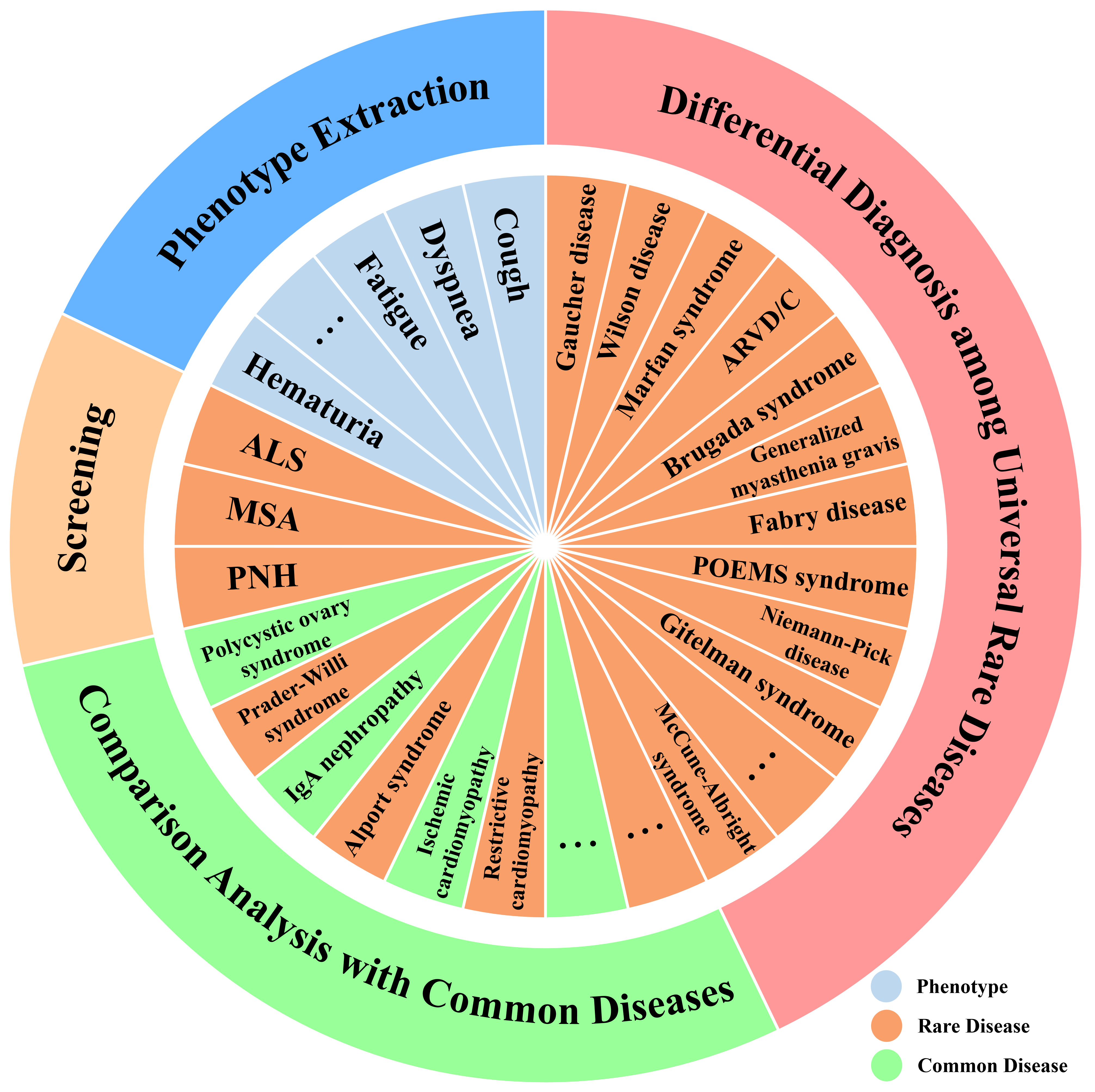}
  \caption{\textit{RareBench}'s overview of evaluation tasks.}
  \Description{}
  \label{fig:overview}
\end{figure}

Our work leverages LLMs to conduct comprehensive evaluations in the challenging field of rare diseases. 
Figure \ref{fig:overview} displays the overview of evaluation dimensions for the four tasks of \textit{RareBench}, and more detailed definitions and descriptions are provided in Section \ref{sec:3}. 
The main contributions are: 
1) \textbf{Dataset and Benchmarking}: We develop a diverse, multi-center, and specifically tailored dataset for rare diseases. 
Alongside this, we introduce \textit{RareBench}, a comprehensive benchmarking framework for evaluating LLMs' capabilities in real-world complex clinical scenarios like phenotype extraction and differential diagnosis.
2) \textbf{Advanced Knowledge Integration Prompt}: We integrate rich knowledge sources to create an exhaustive knowledge graph for rare diseases. 
Utilizing a disease-phenotype graph and the hierarchical structure of the phenotype graph, we devise a novel random walk algorithm capitalizing on phenotype \textit{Information Content} (IC) values to implement dynamic few-shot prompting strategies. 
This advancement significantly boosts the performance of LLMs excluding GPT-4 in differential diagnosis, even surpassing GPT-4.
3) \textbf{Human-versus-LLMs Comparative Studies}: We demonstrate GPT-4 on par with senior doctors across five distinct specialties in the differential diagnosis of rare diseases through comparative analysis. 
The experiments show that GPT-4’s diagnostic capabilities in rare diseases are now commensurate with those of experienced specialist physicians.

\section{Related Work}

\begin{figure*}[ht]
  \includegraphics[width=\textwidth]{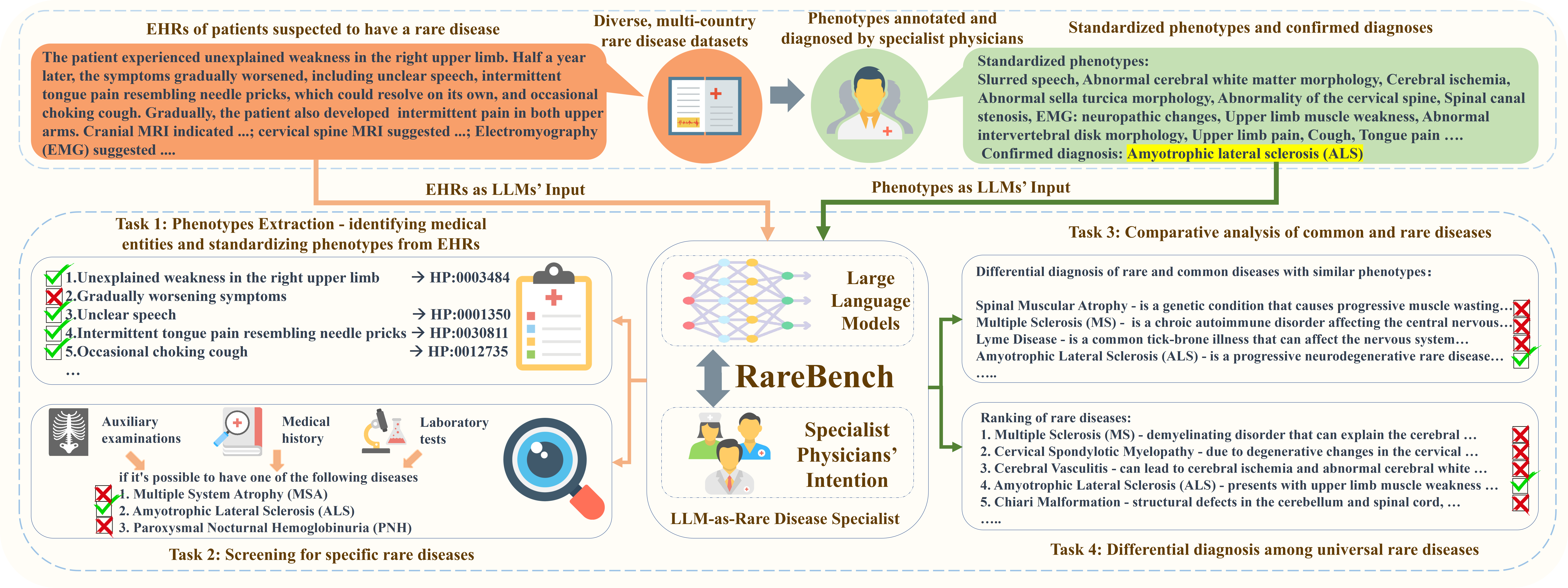}
  \caption{\textit{RareBench} is the first benchmark to evaluate LLMs as rare disease specialists on 4 distinct tasks.}
  \Description{}
  \label{fig:mainpic}
\end{figure*}

\subparagraph{\textbf{Medical Benchmarks for LLMs.}}

Prominent medical question-and-answer data benchmarks, such as MedQA \cite{jin2021disease}, PubMedQA \cite{jin2019pubmedqa}, MedMCQA \cite{pal2022medmcqa}, MultiMedQA \cite{singhal2023large}, and CMExam \cite{liu2023benchmarking}, primarily derive from medical examinations, typically featuring multiple-choice formats. 
MedBench \cite{cai2023medbench} introduces a large-scale Chinese benchmark for evaluating LLMs in clinical knowledge and diagnostics. 
Additionally, DDXPlus \cite{fansi2022ddxplus} provides a detailed medical diagnostic dataset covering symptoms and diagnoses. 
Our \textit{RareBench} extends this landscape by focusing on complex clinical scenarios specific to rare diseases.

\subparagraph{\textbf{LLMs' Medical Capability.}}

The evolution of General Medical Artificial Intelligence (GMAI) is reshaping healthcare by automating learning processes and incorporating domain knowledge to reduce the clinical workload \cite{moor2023foundation}.
GPT-4 \cite{achiam2023gpt}, a notable example in this field, has demonstrated exceptional skills in medical questions \cite{nori2023capabilities} and rivals or surpasses state-of-the-art models in tasks such as radiology text interpretation \cite{liu2023exploring}. 
Medprompt \cite{nori2023can} further enhances this capability through specialized prompting techniques, enabling foundational models like GPT-4 to outperform dedicated healthcare models. 
Besides GPT-4, models like AMIE \cite{tu2024towards} show superior performance in specific medical tasks, even exceeding the diagnostic abilities of general primary care physicians in some cases. 
These LLMs not only assist in differential diagnosis but also engage in clinical reasoning, thus improving diagnostic accuracy \cite{mcduff2023towards, kwon2023large}. 
Moreover, fine-tuned LLMs can efficiently extract valuable data from clinical notes, significantly boosting patient care quality \cite{vaid2023using}. 
Despite these advancements, challenges in accuracy, interpretability, and safety persist in the medical application of LLMs, underscoring the need for continuous refinement \cite{thirunavukarasu2023large, li2023ethics}. 
Notably, the potential of LLMs in rare disease contexts is yet to be fully explored, and our research aims to fill this gap.

\subparagraph{\textbf{Diagnosis of Rare Disease.}}

The initial step in clinical diagnosis involves extracting standardized phenotypes from a patient's electronic health record (EHR). 
To translate clinical texts into standardized Human Phenotype Ontology (HPO) terms, various natural language processing (NLP) methods \cite{son2018deep, liu2019doc2hpo, deisseroth2019clinphen, luo2021phenotagger, feng2022phenobert, yang2023enhancing} have been developed. 
For rare disease diagnosis, current computational methods comprise many statistical or machine learning-based methods \cite{zhai2023phen2disease, robinson2020interpretable, li2019xrare, zhao2020phen2gene, son2018deep, kohler2009clinical, jia2018rdad, pinol2017rare, peng2016measuring}.

\section{Composition of \textit{RareBench}}
\label{sec:3}
This section presents 4 critical tasks of the \textit{RareBench} framework with the most extensive collection of rare disease patient datasets currently accessible. 
These tasks include 1) phenotype extraction from EHRs, 2) screening for specific rare diseases, 3) comparative analysis of common and rare diseases, and 4) differential diagnosis among universal rare diseases. 
Figure \ref{fig:mainpic} demonstrates the process of employing LLMs as rare disease specialists to complete the four tasks in the \textit{RareBench} using an EHR or a set of phenotypes of an ALS (Amyotrophic Lateral Sclerosis) patient from PUMCH.
We also describe our prompting techniques for effectively deploying LLMs as rare disease specialists. 
Please refer to the Appendix due to the page limit for dataset details and prompt examples.

\subsection{Tasks of the \textit{RareBench} Framework}

\subsubsection{Task 1: Phenotype Extraction from Electronic Health Records}

Task 1 involves deriving phenotypes from EHRs for diagnosis. We design 3 sub-tasks: 
a) \textbf{Phenotype Extraction}: Extracting phenotypes from EHRs precisely. 
b) \textbf{General Entity Extraction}: Extracting general entities from EHRs.
c) \textbf{General Entity Standardization}: Standardizing general entities into phenotypes.
\textit{General entity extraction} and \textit{standardization} can be considered a two-step decomposition of \textit{phenotype extraction}, allowing for a more detailed evaluation of LLMs' capabilities of phenotype extraction.

\subsubsection{Task 2: Screening for Specific Rare Diseases}

Task 2 aims to evaluate the capability of LLMs in identifying risk factors or symptoms associated with specific rare diseases. It utilizes patients' medical histories and auxiliary examinations to discover potential rare diseases and facilitate further diagnosis. For this study, three rare diseases are selected: ALS (Amyotrophic Lateral Sclerosis), PNH (Paroxysmal Nocturnal Hemoglobinuria), and MSA (Multiple System Atrophy).

\subsubsection{Task 3: Comparison Analysis of Common and Rare Diseases}

Task 3 aims to validate whether LLMs can differentiate between patients with common and rare diseases that exhibit similar phenotypes/symptoms. From the PUMCH dataset, we select 527 electronic health records containing 60 cases with 13 rare diseases and 467 cases with 64 common diseases, respectively. The task of LLMs is to predict the top ten most likely diseases from the mentioned pool of 77 diseases based on a patient’s electronic health record. 

\subsubsection{Task 4: Differential Diagnosis among Universal Rare Diseases}

Task 4 is centered on a systematic differential diagnosis across \textbf{the full spectrum of known rare diseases} to identify the most probable rare disease. 
In this process, specialist physicians consider a range of possible rare diseases and methodically narrow down the potential diagnoses through a process of elimination or additional diagnostic tests. 
After gathering adequate evidence, they determine the most likely diagnosis. 
Unlike task 3, \textbf{this task does not limit the range of rare diseases}. Instead, it leverages LLMs to provide the top ten most likely diagnoses based solely on the patient's phenotypes/symptoms. 
Task 4 involves 2,185 patient cases encompassing 421 rare diseases, collected from both Public and PUMCH datasets. Its primary objective is to evaluate LLMs' capability within the complexities of real-world clinical scenarios. As the most pivotal and forward-thinking task of \textit{RareBench}, it plays a crucial role in accessing the potential of LLMs in handling intricate medical data.

\begin{table}[h]
    \centering
    \caption{Rare disease datasets of \textit{RareBench}'s four tasks.}
    \footnotesize
    \begin{tabular}{lcccc}
        \hline
        Dataset & Type & \#.Diseases & \#.Cases &Source  \\ 
        \hline
        Task 1   & EHRs & 34 & 87 & PUMCH\\
        Task 2 & EHRs & 3 & 33  & PUMCH\\
        Task 3 & EHRs & 77 & 527 &  PUMCH\\
        Task 4  & EHRs/symptoms & 421 & 2,185 & Public \& PUMCH \\
        \hline
    \end{tabular}
    \label{tab:dataset_description}
\end{table}

\subsection{Rare Disease Patients' Dataset}

This study categorizes datasets into two main groups: publicly available datasets and the Peking Union Medical College Hospital (PUMCH) datasets. All of these data are utilized to perform one of the four tasks, as illustrated in Table \ref{tab:dataset_description}. The publicly available datasets provide phenotypes/symptoms alongside confirmed rare diagnoses in OMIM/Orphanet disease codes. Consequently, they can only be employed to assess the performance of \textit{differential diagnosis among universal rare diseases} (Task 4). In total, public datasets include 1,122 cases spanning over 362 rare diseases. Further descriptions can be found in Appendix \ref{app:dataset}.

Electronic health records from PUMCH serve as valuable resources for both phenotype extraction and three diagnostic tasks. The PUMCH dataset comprises a total of 1,650 cases, consisting of 1,183 rare disease cases and 467 common disease cases. Among them, we select specific datasets for each of the four tasks. Specifically, we first choose 87 EHRs involving 34 rare diseases annotated with physician-identified phenotypes to perform task 1. Next, we select 33 cases with complete auxiliary examination and medical history information to execute task 2. Then, from the remaining medical records (excluding those for tasks 1 and 2), we select 60 cases with rare diseases and 467 cases with common diseases exhibiting phenotypes similar to those of the corresponding rare diseases for task 3. Finally, for task 4, we test all the remaining rare disease records (excluding those for tasks 1 and 2), along with all the aforementioned public cases, totaling 2,185 cases. 
It's important to note that the patient's personally identifiable information has been removed completely. Additionally, doctors from PUMCH monitored all cases before uploading text information, ensuring the absence of any potential personal information leaks. 
Moreover, we apply reasonable filtering criteria to identify and remove cases of low quality that may be caused by recording errors or missing information, such as those with uncertain or imprecise diagnoses and those lacking sufficient relevant information, i.e., fewer than three phenotypes.

\subsection{Framework Setup}

\subsubsection{Evaluated Models}

We select eleven models in our evaluation framework, including API-based and open-source LLMs, as detailed in Table \ref{tab:model_description}. 
Specifically, we choose 5 API-based models, which exhibit superior performance as a result of substantial investment and advanced development. 
On the other hand, due to limitations in computational resources, our selection of open-source models is confined to 3 general LLMs and 3 medical LLMs, each with a model size of fewer than 10 billion parameters. 

\begin{table}[h]
    \centering
    \caption{Eleven LLMs evaluated as rare disease specialists.}
    \begin{tabular}{lccc}
        \hline
        Model & \#Size & Form & Version\\ \hline
        GPT-4 \cite{achiam2023gpt} & N/A & API & 1106-preview \\
        GPT-3.5-Turbo \cite{noauthor_introducing_nodate} & N/A & API & 1106 \\
        Gemini Pro \cite{team2023gemini} & N/A & API & - \\
        GLM4 \cite{zeng2022glm, du2022glm} & N/A & API & -\\
        GLM3-Turbo \cite{zeng2022glm, du2022glm} & N/A & API & -\\ \hline
        Mistral-7B \cite{jiang2023mistral} & 7B & Open Source & instruct-v0.1\\
        Llama2-7B \cite{touvron2023llama} & 7B & Open Source & chat\\
        ChatGLM3-6B \cite{zeng2022glm, du2022glm} & 6B & Open Source & - \\ 
        BioMistral-7B \cite{labrak2024biomistral} & 7B & Open Source & -\\
        HuatuoGPT2-7B \cite{chen2023huatuogptii} & 7B & Open Source & -\\
        MedAlpaca-7B \cite{han2023medalpaca} & 7B & Open Source & - \\
        \hline
    \end{tabular}
    \label{tab:model_description}
\end{table}

\subsubsection{Basic Prompt Design}

For the evaluation of 11 LLMs, we primarily utilize the most fundamental zero-shot prompt. 
We assign the role of a rare diseases specialist to the LLMs by incorporating \textit{"You are a specialist in the field of rare diseases."} as the system prompt/initial statement. 
Additional details on the configuration of LLMs' hyper-parameters are available in Appendix \ref{app:framework}.

\subsubsection{More Prompting Strategies Exploration on GPT-4}

We further explore diverse prompting strategies with GPT-4, including Chain-of-Thought (CoT) \cite{wei2022chain} and random few-shot methods. 
In the CoT settings, the zero-shot prompt is supplemented with \textit{"Let us think step by step."} to foster a sequential thought process, a technique validated in various general tasks. 
For random few shots, we provide LLMs with $m$ random complete input-output examples as prompts, where the choice of $m$ depends on the specific task.

\subsection{Knowledge Integration Dynamic Few-shot}

\begin{figure}
    \centering
    \includegraphics[width=1\linewidth]{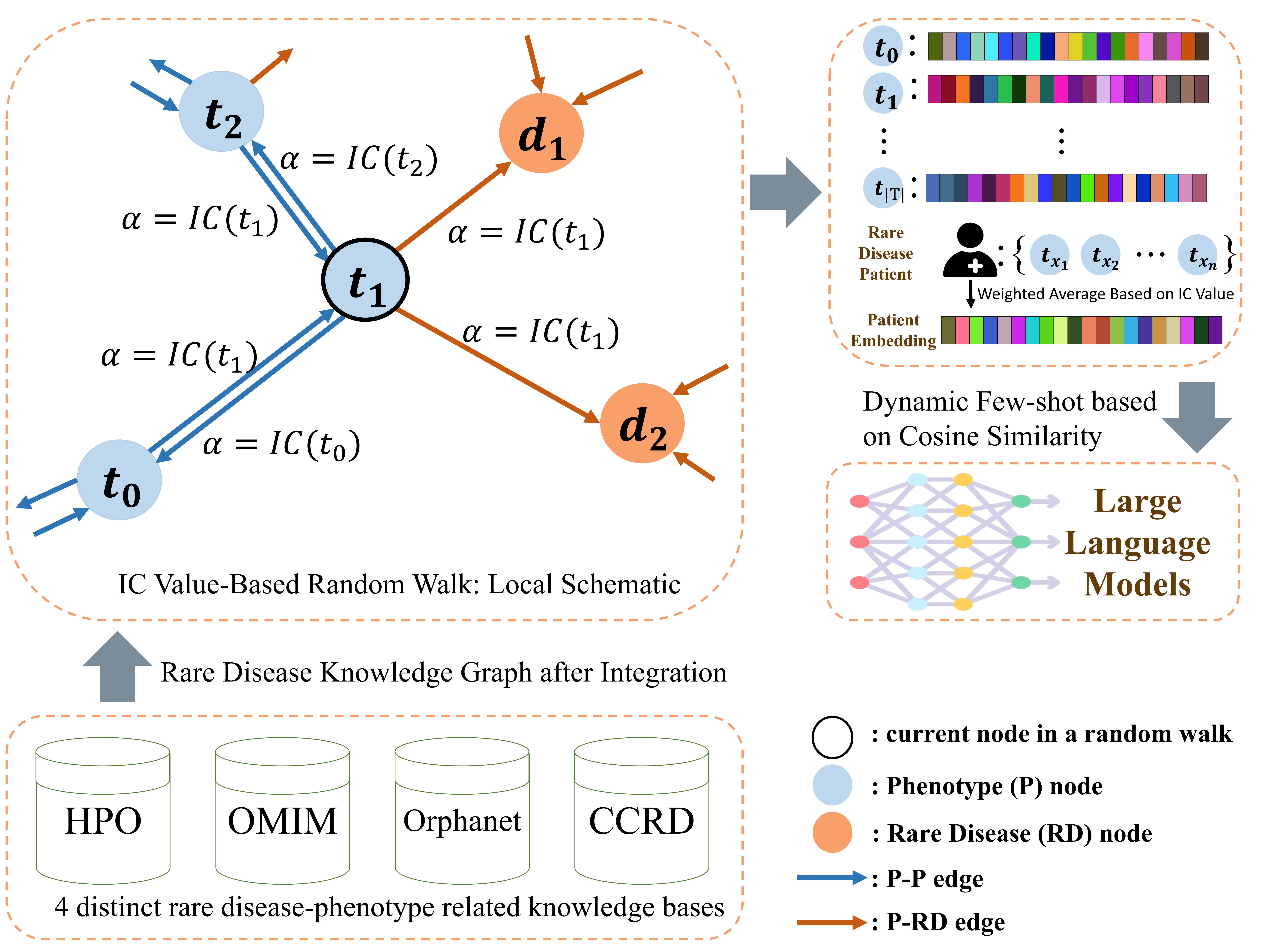}
    \caption{The workflow of the dynamic few-shot strategy includes an integrated rare disease knowledge graph from 4 knowledge bases and an IC value-based random walk algorithm for phenotype and disease embedding.}
    \Description{}
    \label{fig:ic}
\end{figure}

Beyond the prompts above, we construct a rare disease domain knowledge graph by integrating multiple public knowledge bases. This serves as the foundation for our implemented \textit{Information Content} (IC) based random walk algorithm. The phenotype embeddings generated via this algorithm have been pivotal in developing a dynamic few-shot prompt, depicted in Figure \ref{fig:ic}. 
This innovative approach markedly improves the capabilities of LLMs in \textit{differential diagnoses among universal rare diseases} (Task 4). 

\subsubsection{Rare Disease Knowledge Graph Construction}

Previous methods \cite{kohler2009clinical, jia2018rdad, pinol2017rare, peng2016measuring} for the differential diagnosis of rare diseases primarily rely on similarity calculations between diseases using phenotypes in knowledge bases. 
It is feasible to construct a knowledge graph wherein both rare diseases and phenotypes are represented as nodes, connected by their interrelations as edges. 
There are two types of edges. The first is phenotype-phenotype edges obtained from the HPO hierarchy that organizes phenotypes into a directed acyclic graph where each edge connects a more specific phenotype (child) to a broadly defined phenotype (parent). The other is the disease-phenotype information, for which we integrate data from 4 rare-disease-related knowledge bases:  the Human Phenotype Ontology (HPO) \cite{kohler2021human,kohler2017human,kohler2009clinical}, Online Mendelian Inheritance in Man (OMIM) \cite{hamosh2005online}, Orphanet \cite{ayme2003orphanet}, and the Compendium of China’s Rare Diseases (CCRD) \cite{he2019incidence}. 
This integration notably enhances the annotation of associations between rare diseases and phenotypes. 
The statistical information of the integrated knowledge graph is presented in Table \ref{tab:KG_stat}, with detailed descriptions of each knowledge base available in Appendix \ref{app:knowledge}.

\begin{table}
    \scriptsize
    \centering
    \caption{Key statistics of the integrated rare disease knowledge graph encompassing phenotype (P) and rare disease(RD) nodes and two types of edges (P-P and P-RD). The asterisk ("$*$") indicates the consolidation of rare diseases from various knowledge bases into 9,260 unique entities.}
    \begin{tabular}{c|cc|cccc}
        \hline
        Type & Phenotype (P) & Rare Disease (RD) & P-P & P-RD & P-RD & P-RD\\ 
        (Src.) & (HPO) & ($*$) & (HPO) & (OMIM) & (Orphanet) & (CCRD)\\\hline
        Num. & 17232 & 9260 & 21505 & 54169 & 98031 & 4223 \\ \hline
    \end{tabular}
    \label{tab:KG_stat}
\end{table}

\subsubsection{Random Walk Based on Information Content}

The concept of \textit{Information Content} (IC) \cite{cover1999elements} is similar to \textit{Inverse Document Frequency} (IDF) utilized in natural language processing. 
IC is employed as an index of a phenotype's specificity to a particular disease. 
Notably, a phenotype's proximity to the root node in the HPO hierarchy (a broad phenotype), or a higher frequency of its association with multiple diseases (a common phenotype), results in a lower IC, reflecting lesser significance. 
On the other hand, a phenotype that is highly specific to a rare disease has a high IC. 
The essence of IC lies in its inverse correlation with the prevalence of a phenotype – the more common a phenotype, the lower its IC. 
Let \( T \) be the complete set of phenotype terms. 
For a given term \( t \in T \), the computation of \( IC(t) \) is formulated as follows: 
$$
IC(t) = -log\frac{n(t)}{N},
$$
where \( n(t) \) represents the count of diseases annotated with the HPO phenotype \( t \) or its descendant phenotypes, and \( N \) signifies the total number of annotated diseases. 
In our integrated rare disease knowledge graph, the value of \( N \) equals 9,260.

Node2vec \cite{grover2016node2vec}, a typical shallow embedding method, is calculated by simulating random walks \cite{perozzi2014deepwalk} across a graph through a flexible and parameterized walk strategy. 
These unsupervised walks generate sequences of nodes, which are employed to create node embeddings by using the methodologies developed for Word2vec \cite{mikolov2013distributed}. 
In Node2vec, the search bias parameter \(\alpha\) controls the search strategy in generating random walk sequences of nodes, effectively modulating the preference for breadth-first search (BFS) or depth-first search (DFS) strategies in exploring neighboring nodes. 
However, a direct application of Node2vec to our phenotype-disease knowledge graph in the rare disease domain falls short of adequately capturing each phenotype’s distinct influence on the differential diagnosis of diseases. 
We innovatively integrated IC values into the Node2vec framework to address this limitation, formulating an IC value-based random walk algorithm. 
This enhancement is designed to enrich the interactions between phenotypes and rare diseases. 
Under this new scheme, when a random walk progresses to a phenotype node \( t_1 \) and the subsequent node is another phenotype \( t_2 \), the walk search bias from \( t_1 \) to \( t_2 \) is determined as \(\alpha = IC(t_2) \). 
Conversely, if the following node is a rare disease \( d_1 \), the search bias from \( t_1 \) to \( d_1 \) is set by \( \alpha = IC(t_1) \). 
This modification ensures that phenotype nodes with higher IC values receive increased focus during the random walk, amplifying associations with the related rare disease nodes.  

\subsubsection{Dynamic Few-shot Prompting Strategy}

Employing the IC value-based random walk algorithm, we developed a function \( f: T \rightarrow \mathbb{R}^d \) to project phenotype terms into a \( d \)-dimensional vector space, where \( d = 256 \). 
These embeddings are then utilized to represent patients with rare diseases.
A patient \( p \) with a rare disease \( d \) is characterized with a set of phenotype terms \( t_{x_i} \), expressed as \( p = \{t_{x_1}, t_{x_2}, ..., t_{x_n}\} \). The embedding of the patient \( p \) is computed as follows:

$$
Embedding(p) = \frac{1}{\sum_{i=1}^{n} IC(t_{x_i})} \sum_{i=1}^{n} IC(t_{x_i}) \cdot f(t_{x_i}).
$$

\begin{table*}[t]
\caption{Comprehensive results of the \textit{RareBench}'s 4 tasks. In task 1, "PTE", "GEE", and "GES" respectively stand for \textit{phenotype extraction}, \textit{general entity extraction}, and \textit{general entity standardization}.\textbf{ Bold numbers} indicate the best results, while \underline{underlined numbers} signify the second-best results.}
\begin{tabular}{l|ccc|ccc|cccc|cccc}
\hline
\multicolumn{1}{c|}{\multirow{4}{*}{LLMs}} & \multicolumn{3}{c|}{Task 1: Phenotype}                        & \multicolumn{3}{c|}{Task 2: Screening}                                                                                              & \multicolumn{4}{c|}{Task 3: Comparative}                                                                                         & \multicolumn{4}{c}{Task 4: Differential Diagnosis}                                                          \\  
\multicolumn{1}{c|}{}                      & \multicolumn{3}{c|}{Extraction from EHRs}                        & \multicolumn{3}{c|}{for Specific RDs}                                                                                              & \multicolumn{4}{c|}{Analysis of Common \& RDs}                                                                                         & \multicolumn{4}{c}{among Universal RDs on 2,185 cases}                                                          \\ \cline{2-15}
\multicolumn{1}{c}{}                      & \multicolumn{3}{|c|}{F1-score (\%) ($\uparrow$)}                                 & \multicolumn{3}{c|}{Recall (\%) ($\uparrow$)}                                                    & \multicolumn{3}{c}{Top-k Recall (\%) ($\uparrow$)}                                                  & \multicolumn{1}{c|}{\multirow{2}{*}{MR ($\downarrow$)}} & \multicolumn{3}{c}{Top-k Recall (\%) ($\uparrow$)}                                                     & \multirow{2}{*}{MR ($\downarrow$)}   \\ 
\multicolumn{1}{c|}{}                      & \multicolumn{1}{c}{PTE}  & \multicolumn{1}{c}{GEE}  & GES                            & \multicolumn{1}{c}{ALS} & \multicolumn{1}{c}{PNH} & \multicolumn{1}{c|}{MSA} & \multicolumn{1}{c}{k=1} & \multicolumn{1}{c}{k=3} & \multicolumn{1}{c}{k=10} & \multicolumn{1}{c|}{}                    & \multicolumn{1}{c}{k=1} & \multicolumn{1}{c}{k=3} & \multicolumn{1}{c}{k=10} &                       \\ \hline
GPT-4 0-shot                                & \multicolumn{1}{c}{24.5} & \multicolumn{1}{c}{\underline{64.9}} & 38.7                           & \underline{62.5}    & \underline{57.1}   & \multicolumn{1}{c|}{\underline{61.1}}                   & \multicolumn{1}{c}{\underline{46.1}}     & \multicolumn{1}{c}{\underline{59.6}}     & \multicolumn{1}{c}{\underline{72.1}}      &  \multicolumn{1}{c|}{\underline{3.0}}                                          & \multicolumn{1}{c}{\underline{32.3}}  & \multicolumn{1}{c}{\underline{45.4}}  & \multicolumn{1}{c}{\underline{58.9}}   & \underline{5.0}                   \\ 
\hspace{9mm}3-shot                                     & \multicolumn{1}{c}{\textbf{26.0}} & \multicolumn{1}{c}{61.9} & \textbf{42.0}                           & \multicolumn{1}{c}{-}    & \multicolumn{1}{c}{-}    &     \multicolumn{1}{c|}{-}                     & \multicolumn{1}{c}{-}     & \multicolumn{1}{c}{-}     & \multicolumn{1}{c}{-}      &      -                                    & \multicolumn{1}{c}{30.4}  & \multicolumn{1}{c}{43.8}  & \multicolumn{1}{c}{57.9}   & \underline{5.0}                   \\ 
\hspace{9mm}CoT                                        & \multicolumn{1}{c}{\underline{25.1}} & \multicolumn{1}{c}{\textbf{65.9}} & \underline{39.7}                           & \textbf{62.5}    & \textbf{57.1}    &      \textbf{66.7}                  & \textbf{47.4}     & \textbf{62.0}     & \textbf{75.0}      &     \multicolumn{1}{c|}{\textbf{2.0}}                                      & \multicolumn{1}{c}{\textbf{33.2}}  & \multicolumn{1}{c}{\textbf{46.4}}  & \multicolumn{1}{c}{\textbf{59.5}}   & \textbf{4.0}                   \\
GPT-3.5-Turbo                               & \multicolumn{1}{c}{17.2} & \multicolumn{1}{c}{48.4} & 32.3                           & \multicolumn{1}{c}{37.5}    & \multicolumn{1}{c}{42.9}    &    \multicolumn{1}{c|}{44.4}                      & \multicolumn{1}{c}{33.2}     & \multicolumn{1}{c}{45.4}     & \multicolumn{1}{c}{58.1}      &            \multicolumn{1}{c|}{5.0}                               & \multicolumn{1}{c}{21.1}  & \multicolumn{1}{c}{34.2}  & \multicolumn{1}{c}{48.2}   & \textgreater{}10      \\ 
Gemini Pro                                  & \multicolumn{1}{c}{10.1} & \multicolumn{1}{c}{50.8} & 34.3                           & \multicolumn{1}{c}{25.0}    & \multicolumn{1}{c}{28.6}    &          \multicolumn{1}{c|}{33.3}                & \multicolumn{1}{c}{24.3}     & \multicolumn{1}{c}{32.6}     & \multicolumn{1}{c}{44.8}      &      \textgreater{}10                                    & \multicolumn{1}{c}{14.6}  & \multicolumn{1}{c}{22.7}  & \multicolumn{1}{c}{33.0}   & \textgreater{}10      \\
GLM4                                        & \multicolumn{1}{c}{15.9} & \multicolumn{1}{c}{56.8} & 24.0                           & \multicolumn{1}{c}{50.0}    & \multicolumn{1}{c}{42.9}    &    \multicolumn{1}{c|}{50.0}                       & \multicolumn{1}{c}{31.3}     & \multicolumn{1}{c}{42.5}     & \multicolumn{1}{c}{57.1}      &              8.0                            & \multicolumn{1}{c}{19.1}  & \multicolumn{1}{c}{30.4}  & \multicolumn{1}{c}{45.5}   & \textgreater{}10      \\
GLM3-Turbo                                  & \multicolumn{1}{c}{12.9} & \multicolumn{1}{c}{53.5} & 31.4                           & \multicolumn{1}{c}{12.5}    & \multicolumn{1}{c}{14.3}    &     \multicolumn{1}{c|}{22.2}                     & \multicolumn{1}{c}{25.8}     & \multicolumn{1}{c}{37.8}     & \multicolumn{1}{c}{51.2}      &        10.0                                  & \multicolumn{1}{c}{12.4}  & \multicolumn{1}{c}{21.0}  & \multicolumn{1}{c}{33.2}   & \textgreater{}10      \\ \hline
Mistral-7B                                  & \multicolumn{1}{c}{3.3}  & \multicolumn{1}{c}{26.4} & 8.8                            & \multicolumn{1}{c}{0.0}    & \multicolumn{1}{c}{14.3}    &  \multicolumn{1}{c|}{11.1}                        & \multicolumn{1}{c}{13.7}     & \multicolumn{1}{c}{19.2}     & \multicolumn{1}{c}{29.0}      &   \textgreater{}10                                        & \multicolumn{1}{c}{7.2}   & \multicolumn{1}{c}{12.5}  & \multicolumn{1}{c}{18.9}   & \textgreater{}10      \\ 
Llama2-7B                                   & \multicolumn{1}{c}{0.0}  & \multicolumn{1}{c}{0.0}  & 0.0                            & \multicolumn{1}{c}{0.0}    & \multicolumn{1}{c}{14.3}    &     \multicolumn{1}{c|}{5.6}                      & \multicolumn{1}{c}{-}     & \multicolumn{1}{c}{-}     & \multicolumn{1}{c}{-}      &    -                                      & \multicolumn{1}{c}{7.4}   & \multicolumn{1}{c}{11.0}  & \multicolumn{1}{c}{14.8}   & \textgreater{}10      \\ 
ChatGLM3-6B                                 & \multicolumn{1}{c}{10.3} & \multicolumn{1}{c}{48.3} & 14.6                           & \multicolumn{1}{c}{12.5}    & \multicolumn{1}{c}{0.0}    &      \multicolumn{1}{c|}{5.6}                    & \multicolumn{1}{c}{9.7}     & \multicolumn{1}{c}{16.1}     & \multicolumn{1}{c}{22.8}      &                \textgreater{}10                           & \multicolumn{1}{c}{5.0}   & \multicolumn{1}{c}{7.2}   & \multicolumn{1}{c}{10.7}   & \textgreater{}10      \\ 
BioMistral-7B                                  & \multicolumn{1}{c}{0.4}  & \multicolumn{1}{c}{14.0} & 5.2                            & \multicolumn{1}{c}{12.5}    & \multicolumn{1}{c}{14.3}    &  \multicolumn{1}{c|}{5.6}                        & \multicolumn{1}{c}{16.7}     & \multicolumn{1}{c}{19.5}     & \multicolumn{1}{c}{24.9}      &   \textgreater{}10                                        & \multicolumn{1}{c}{6.5}   & \multicolumn{1}{c}{8.5}  & \multicolumn{1}{c}{12.3}   & \textgreater{}10      \\ 
HuatuoGPT2-7B                                  & \multicolumn{1}{c}{3.7}  & \multicolumn{1}{c}{17.8}  & 6.6                            & \multicolumn{1}{c}{12.5}    & \multicolumn{1}{c}{28.6}    &     \multicolumn{1}{c|}{11.1}                      & \multicolumn{1}{c}{18.6}     & \multicolumn{1}{c}{30.4}     & \multicolumn{1}{c}{40.8}      &    \textgreater{}10                                      & \multicolumn{1}{c}{11.4}   & \multicolumn{1}{c}{17.8}  & \multicolumn{1}{c}{28.1}   & \textgreater{}10      \\ 
MedAlpaca-7B                                 & \multicolumn{1}{c}{0.0} & \multicolumn{1}{c}{0.0} & 0.0                          & \multicolumn{1}{c}{0.0}    & \multicolumn{1}{c}{14.3}    &      \multicolumn{1}{c|}{0.0}                    & \multicolumn{1}{c}{-}     & \multicolumn{1}{c}{-}     & \multicolumn{1}{c}{-}      &                -                           & \multicolumn{1}{c}{8.4}   & \multicolumn{1}{c}{14.3}   & \multicolumn{1}{c}{19.4}   & \textgreater{}10 \\
\hline 

\end{tabular}
\label{tab:main_result}
\end{table*}

In MedPrompt \cite{nori2023can}, the dynamic few-shot method selects training examples that are most similar to the specific input. 
However, it relies on a general-purpose text embedding model such as \textit{text-embedding-ada-002} \cite{openai_embedding}, which can only measure the relatedness of phenotype text, without considering deeper associations among phenotypes. 
Therefore, we utilize the phenotype embeddings generated from our IC value-based random walk algorithm to represent rare disease patients. 
We then select the top $m$ most similar examples from the rare disease patient datasets. Data with the highest cosine similarity serve as prompts. 
From the LLMs' perspective, such a retrieval augmented generation (RAG) \cite{lewis2020retrieval} strategy enables LLMs to be more effectively tailored to the rare disease domain by conforming more closely to differential diagnosis tasks and minimizing hallucinations through enriched knowledge. 
From a clinical perspective, the diagnostic process of specialist physicians relies on medical knowledge and past experiences of those previously diagnosed patients. 
Consequently, a curated set of relevant examples grants LLMs a distilled version of "clinical experience". 

It should be noted that there are many other embedding models available that can be used for choosing examples for LLM’s dynamic few-shot prompts, such as MedPrompt \cite{nori2023can} and Auto-CoT \cite{zhang2022automatic}. However, our IC-value-based random walk strategy is a simple, easy-to-implement method that captures the critical concept of differential diagnosis. Its effectiveness will be evaluated against other state-of-the-art few-shot methods.

\section{Evaluation Results of RareBench}

This section presents the comprehensive results of \textit{RareBench} in Table \ref{tab:main_result}, including the evaluation of GPT-4 across three different settings: zero-shot, few-shot, and Chain of Thought (CoT) with the zero-shot performance of the other 10 LLMs.

\subsection{Task 1: Phenotype Extraction from EHR}

\subsubsection{Metric}

Task 1 is evaluated using precision, recall, and \textbf{F1-score}. Accuracy requires exact matches with the reference for \textit{phenotype extraction} and \textit{general entity standardization}. For \textit{general entity extraction}, predictions are correct if they convey the same meaning as the reference.

\subsubsection{Results}
 
Although GPT-4 achieves the best among all LLMs, the results show that all LLMs perform poorly in \textit{phenotype extraction}. 
The \textit{general entity extraction} and \textit{standardization} results show that LLMs perform well in entity extraction but have weak capabilities in standardizing general entities into phenotypes. This suggests that the main reason for the poor performance of LLMs in \textit{phenotype extraction} is their weak standardization ability. 

Beyond Chain-of-Thought (CoT) and random few-shot, our experimentation includes two additional methods: a) Word-to-phenotype matching, where each word in the output is aligned with the semantically nearest word in the HPO phenotype list. The semantic distance is measured by vectorizing words using OpenAI's \textit{text-embedding-ada-002} \cite{openai_embedding} model and then assessing cosine similarity; 2) Expanded semantic matching, where we associate each output with the semantically closest $n$ words in the HPO phenotype list, then integrate these matches as a reference output range before re-querying GPT-4. In our tests, $n$ is set at 20. 
The F1-score of these two experiments is 0.306 and 0.322, respectively.

\subsection{Task 2: Screening for Specific Rare Diseases}

\subsubsection{Metric}

Screening for three specific rare diseases (ALS, PNH, MSA) is measured using \textbf{recall}.

\subsubsection{Results}

For this task, GPT-4 again achieves the best performance on all three diseases, with recalls exceeding 0.55 in both zero-shot and Chain of Thought (CoT) settings, respectively. Meanwhile, we did not conduct few-shot learning on GPT-4 due to the limited number of similar cases, as we were concerned that it might influence or bias the subsequent diagnostic results. Other API-based LLMs have recalls of less than or equal to 0.50 on all three diseases. Open-source LLMs have top-1 recalls of less than 0.30 on all three diseases. The use of the CoT approach by GPT-4 results in a slight performance improvement. Overall, LLMs demonstrate the potential for screening three rare diseases using patient information, such as medical history, auxiliary examinations, and laboratory tests.

\subsection{Task 3: Comparison Analysis of Common \& Rare Diseases}

\subsubsection{Metric}

Task 3 employs two key metrics: \textbf{top-k recall (hit@k, where k=1, 3, 10)} and \textbf{median rank (MR)}. 
Top-k recall evaluates the diagnostic accuracy, deeming a diagnosis correct if the actual disease appears within the top-k predictions. 
Median rank represents the median position of accurate diagnoses within predictions across all cases.

\subsubsection{Results}

In this task, GPT-4 achieves a top-1 recall of 0.461 under 0-shot settings (with top-3 and top-10 recalls being 0.596 and 0.721, respectively) and a median rank of 3.0. In comparison, under the Chain of Thought (CoT) setting, GPT-4 achieves the best performance on all metrics. The CoT approach contributes to a moderate improvement in GPT-4's performance. Additionally, we did not perform few-shot learning on GPT-4 due to the limited number of cases, as we were concerned that it might influence or bias the subsequent diagnostic results.
Furthermore, the GPT3.5-Turbo achieves a top-1 recall of 0.332, a top-3 recall of 0.454, a top-10 recall of 0.581, and a median rank 5.0, achieving the second-best performance after the GPT-4 method.
The third-best performance is exhibited by GLM4, with a top-1 recall of 0.313, a top-3 recall of 0.425, a top-10 recall of 0.571, and a median rank 8.0. The other two API-based models have top-1 recalls less than 0.26, with median ranks greater than 10.0. All the open-source models yield top-1 recalls less than 0.20, and all have median ranks greater than 10.0. Additionally, we did not present the results of Llama2-7B because we found its performance on lengthy Chinese EHR texts too poor to output normal results.

\subsection{Task 4: Differential Diagnosis among Universal Rare Diseases}

\subsubsection{Metric}

Task 4 employs the same metric as Task 3.

\subsubsection{Results}

We yield the following notable findings in our extensive dataset comprising 2,185 patients with 421 distinct rare diseases. 
For differential diagnosis of universal rare diseases, GPT-4 achieves a top-1 recall of 0.323 under 0-shot settings (with top-3 and top-10 recalls being 0.454 and 0.589, respectively) and a median rank of 5.0. 
Interestingly, GPT-4's performance slightly decreases under random 3-shot, while adopting the Chain of Thought (CoT) approach leads to a performance improvement with a median rank of 4.0. 
In contrast, the top-1 recalls for the other 4 API-based LLMs are around 0.2, with all their median ranks exceeding 10.0; the top-1 recalls for all 6 LLMs with fewer than 10 billion parameters are less than 0.12. Notably, HuatuoGPT2-7B outperforms models like Mistral-7B and Llama2-7B but falls short against API-based LLMs like GPT-4. Conversely, BioMistral-7B's performance drops post-training on general biomedical data, indicating that crafting or refining LLMs for the rare disease sector could be a fruitful future endeavor. 
In conclusion, GPT-4 demonstrated promising differential diagnosis results in a dataset of 2,185 rare disease patients, encompassing 421 different rare diseases. For a comparison of GPT-4 with human physicians, see Section \ref{sec:human}.

\section{Analysis and Discussion}

This section provides a detailed explanation of the performance enhancement for LLMs in Task 4 (\textit{differential diagnosis among universal rare diseases}) using the dynamic few-shot prompt method based on the knowledge graph. Additionally, it includes a comparison and thorough analysis and discussion of the diagnostic capabilities in rare diseases between doctors from PUMCH and LLMs, using a high-quality subset of the PUMCH dataset.

\subsection{Knowledge Integration Dynamic Few-shot}

\begin{figure}[h]
    \includegraphics[width=1\columnwidth]{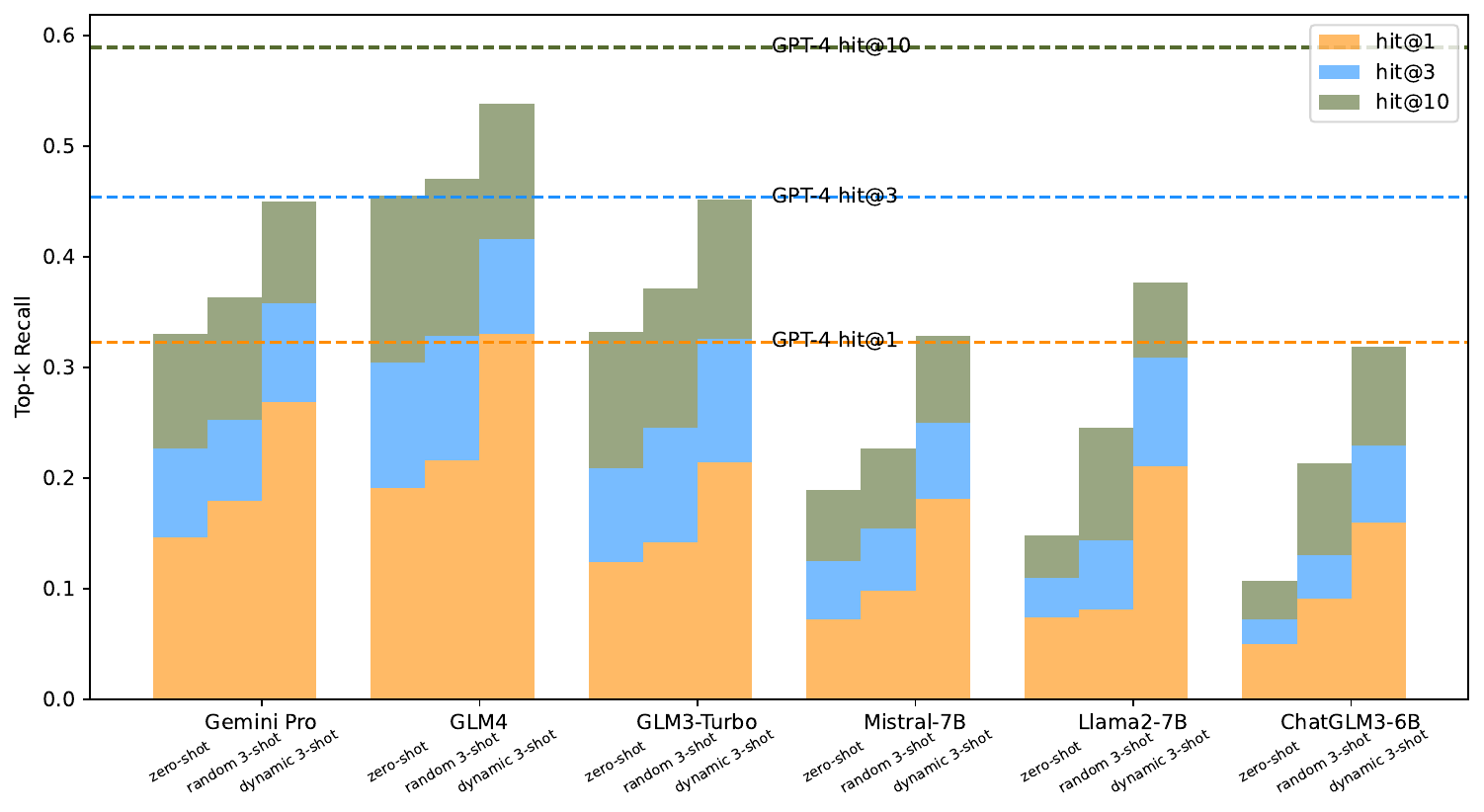}
    \caption{Performance of six LLMs in rare disease differential diagnosis under zero-shot, random 3-shot, and dynamic 3-shot prompts (using GPT-4 zero-shot as a baseline).}
    \Description{}
    \label{fig:IC_dynamic}
\end{figure}

In our research, the IC value-based random walk algorithm we developed is employed to produce embeddings for Human Phenotype Ontology (HPO) phenotype nodes within our comprehensively integrated rare disease knowledge graph. These embeddings are subsequently utilized in dynamic few-shot settings. 
Our experimental evaluation, involving 2,185 rare disease patients and conducted on six LLMs, including GLM4 \cite{du2022glm, zeng2022glm}, is benchmarked against the zero-shot performance of GPT-4. 
The results are illustrated in Figure \ref{fig:IC_dynamic}, with key findings summarized below:

\begin{table*}[ht]
\footnotesize
\tabcolsep=0.35mm
\centering
\caption{Comparison of various prompts' performance across 6 LLMs on 75 high-quality patient cases from PUMCH.}
\begin{tabular}{l|cccc|cccc|cccc|cccc|cccc|cccc}
\hline
\textbf{LLM} & \multicolumn{4}{c|}{\textbf{Gemini Pro}} & \multicolumn{4}{c|}{\textbf{GLM4}} & \multicolumn{4}{c|}{\textbf{GLM3-Turbo}} & \multicolumn{4}{c|}{\textbf{Mistral-7B}} & \multicolumn{4}{c|}{\textbf{Llama2-7B}} & \multicolumn{4}{c}{\textbf{ChatGLM3-6B}} \\

Prompt & hit@1 & hit@3 & hit@10 & MR & hit@1 & hit@3 & hit@10 & MR & hit@1 & hit@3 & hit@10 & MR & hit@1 & hit@3 & hit@10 & MR & hit@1 & hit@3 & hit@10 & MR & hit@1 & hit@3 & hit@10 & MR \\
\hline
zero-shot & 36.0 & 56.0 & 65.3 & 2.0 & 37.3 & 53.3 & 62.7 & 3.0 & 21.3 & 41.3 & 60.0 & 6.0 & 16.0 & 24.0 & 36.0 & >10 & 21.3 & 37.3 & 49.4 & >10 & 9.3 & 16.0 & 22.7 & >10 \\
random 3-shot & 37.3 & 53.3 & 73.3 & 3.0 & 40.0 & 52.0 & 65.3 & 2.0 & 25.3 & 41.3 & 54.7 & 7.0 & 9.3 & 18.7 & 33.3 & >10 & 16.0 & 25.3 & 38.7 & >10 & 9.3 & 13.3 & 17.3 & >10 \\
MedPrompt (3-shot) & 56.0 & 66.7 & 73.3 & 1.0 & 52.0 & 64.0 & 72.0 & 1.0 & 40.0 & 60.0 & 74.7 & 3.0 & 38.7 & 52.0 & 52.0 & 3.0 & 61.3 & \textbf{78.7} & \textbf{81.3} & 1.0 & 30.7 & \textbf{42.7} & \textbf{49.3} & >10 \\
Auto-CoT (3 clusters) & 40.0 & 61.3 & 76.0 & 2.0 & 36.0 & 50.7 & 66.7 & 3.0 & 36.0 & 42.7 & 54.7 & 6.0 & 13.3 & 30.7 & 38.7 & >10 & 20.0 & 30.7 & 40.0 & >10 & 9.3 & 12.0 & 24.0 & >10 \\
dynamic 3-shot (ours) & \textbf{62.7} & \textbf{74.7} & \textbf{82.7} & \textbf{1.0} & \textbf{62.7} & \textbf{70.7} & \textbf{76.0} & \textbf{1.0} & \textbf{72.0} & \textbf{77.3} & \textbf{84.0} & \textbf{1.0} & \textbf{38.7} & \textbf{52.0} & \textbf{64.0} & \textbf{3.0} & \textbf{66.7} & \underline{73.3} & \underline{80.0} & \textbf{1.0} & \textbf{34.7} & \underline{41.3} & \underline{48.0} & >10 \\
\hline
\end{tabular}
\label{tab:new_prompt}
\end{table*}

\begin{itemize}[leftmargin=*]
    \item A holistic analysis reveals that dynamic 3-shot settings significantly enhance the performance of the six LLMs compared to their zero-shot capabilities. On average, there is a notable \textbf{108\% increase in top-1 recall} and substantial improvements of 80\% and 58\% in top-3 and top-10 recalls, respectively.
    \item With dynamic 3-shot, GLM4 outperforms GPT-4's 0-shot level slightly in top-1 recall. In addition, models with less than 10 billion parameters, like Llama2-7b, achieve 0-shot performance on par with more advanced models, such as Gemini Pro and GLM4.
    \item Our ablation study demonstrates an average enhancement of 23\% in top-1 recall across the six LLMs under random 3-shot settings, corresponding increases of 20\% and 21\% in top-3 and top-10 recalls, respectively. This highlights the effectiveness of our knowledge graph-based dynamic few-shot approach in boosting the capabilities of LLMs, particularly those far behind GPT-4, in the context of rare disease differential diagnosis.
\end{itemize}

We also compared our approach against MedPrompt \cite{nori2023can} and Auto-CoT \cite{zhang2022automatic} using 75 high-quality patient cases from PUMCH (employed in our Human-versus-LLMs experiment below) across 6 LLMs. Detailed results are shown in the Table \ref{tab:new_prompt}. Our Dynamic Few-shot shows superior or comparable performance, notably exceeding GPT-4 and specialist physicians in some cases. However, Auto-CoT underperforms with smaller LLMs like Llama2-7b, highlighting the importance of rationale quality in diagnostics. 

\subsection{Human versus LLMs experiment on a subset of Task 4}
\label{sec:human}
\subsubsection{Selection of Test case and Clinical Specialist}
We selected a subset from the PUMCH dataset to compare the diagnostic performance of physicians with various LLMs. The selection criteria included a wide range of diseases across human organ systems, and multiple cases were included for each disease to assess the difficulty level in diagnosis. Ultimately, this test set comprised 75 cases spanning 16 diseases across 5 hospital departments, including Cardiology, Hematology, Nephrology, Neurology, and Pediatrics. To ensure the integrity of the diagnostic process, any information within the EHRs that could potentially provide clues about the underlying diseases was removed.

A total of 50 physicians were employed from 23 Class-A tertiary hospitals in China. Each of the 5 departments had ten physicians. To minimize diagnostic result errors, we assigned each case to be diagnosed by 4 physicians. In the diagnostic process, clinical specialists rely initially on their knowledge and experience to diagnose cases. For situations where the diagnosis is uncertain, clinical specialists can use external assistance such as books and web tools.

\subsubsection{Results}
We initially investigated two input approaches in our comparative experiments. One involved a feature-based input, where the patient's personal phenotype information was provided to inquire about LLMs' diagnostic results. The other approach involved entering the patient's EHR text information with all personally identifiable information removed. We compared the diagnostic outcomes of two input forms on GPT-4 using the test set from the aforementioned 75 PUMCH cases. In contrast, specialist physicians rely solely on EHR text for diagnosis, as shown in Table \ref{tab:ehr_or_phenotype}. When utilizing phenotype input, GPT-4 achieves a top-1 recall of 0.520, a top-3 recall of 0.747, a top-10 recall of 0.827, and a median rank of 1.0. Conversely, EHR text input with GPT-4 results in a top-1 recall of 0.453, a top-3 recall of 0.693, a top-10 recall of 0.800, and a median rank of 2.0. Using extracted phenotypes significantly reduces the number of input tokens for LLMs and marginally outperforms the direct use of medical record text in differential diagnosis. This enhancement stems from the extracted phenotypes summarizing crucial information about the patient's symptoms while eliminating irrelevant details, rendering this input format more economical and efficient for LLMs. In contrast, physicians' diagnostic performance enhancement when aided by external assistance. The top-1 recall increases from 0.407 to 0.447, the top-3 recall rises from 0.468 to 0.511, and the top-10 recall increases from 0.481 to 0.524.

\begin{table}
    \small
    \tabcolsep=0.8mm
    \centering
    \caption{The differential diagnosis performance of GPT-4 with two different input forms and 50 physicians with and without assistance on 75 PUMCH cases.}
    \begin{tabular}{c|cccc}
    \hline
         & hit@1(\%) & hit@3(\%) & hit@10(\%) & MR ($\downarrow$) \\ \hline
        GPT-4 (Phenotypes) & \textbf{52.0} & \textbf{74.7} & \textbf{82.7} & \textbf{1.0} \\
        GPT-4 (EHR text) & 45.3 & 69.3 & 80.0 & 2.0 \\ \hline
        Physicians w/o assistance & 40.7 & 46.8 & 48.1 & - \\
        Physicians w/ assistance & 44.7 & 51.1 & 52.4 & - \\ \hline
    \end{tabular}
    \label{tab:ehr_or_phenotype}
\end{table}

\begin{figure}[h]
    \centering
    \includegraphics[width=1\linewidth]{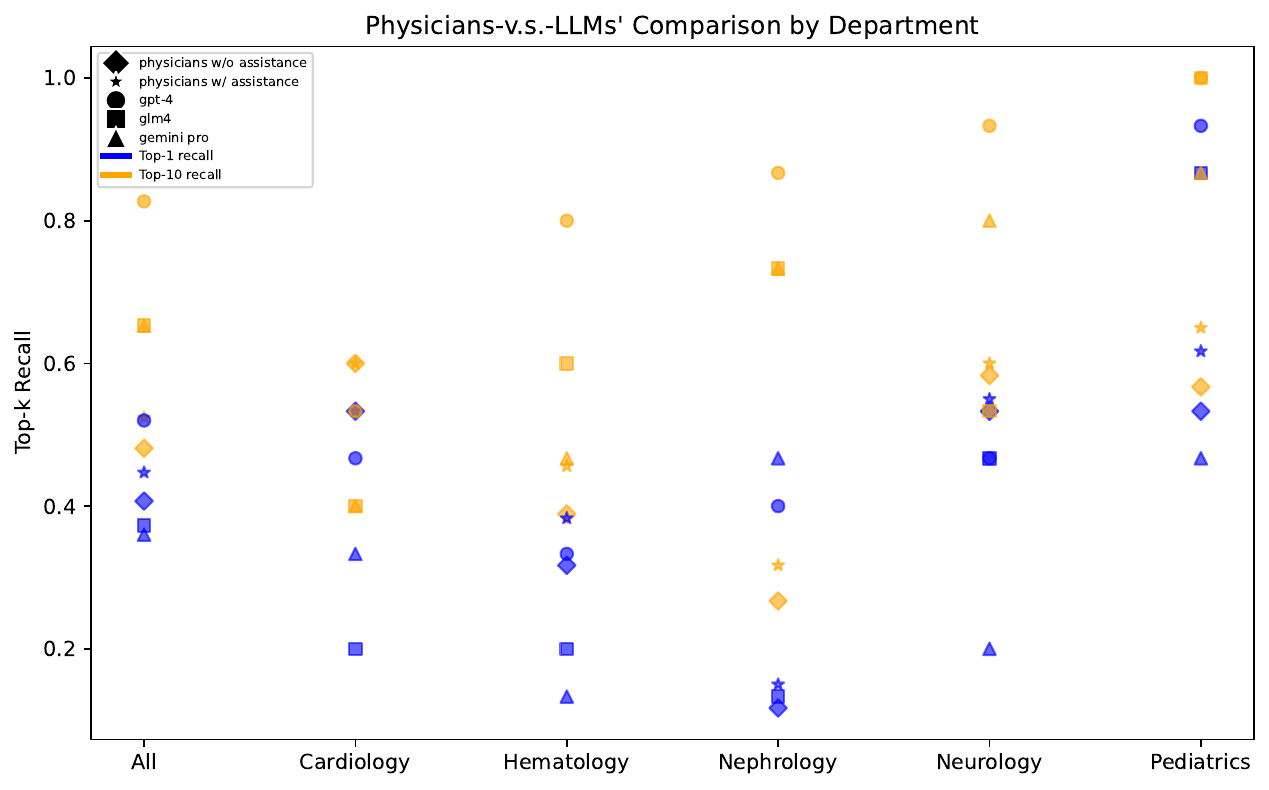}
    \caption{Comparison of top-1 and top-10 recalls in rare disease differential diagnosis across 5 departments between specialist physicians with/without assistance and three LLMs.}
    \Description{}
    \label{fig:pk}
\end{figure}

We also provide a more detailed presentation of GPT-4's performance across diseases related to different medical departments. 
As shown in Figure \ref{fig:pk}, GPT-4 achieved the best diagnostic results across all 5 departments, surpassing the level of specialist physicians. In contrast, the diagnostic performance of the other two LLMs was inferior to that of specialist physicians. Moreover, there were significant differences in diagnostic results among different departments. For example, all methods achieved high recall rates in pediatrics, possibly due to the comprehensive information available in pediatric cases. Conversely, all methods performed poorly in the cases from the Department of Cardiology, which was speculated to be associated with the inherent features of the diseases themselves. Heart diseases often exhibit similarities and overlaps in symptoms and signs, such as chest pain, chest tightness, palpitations, edema, etc. The differential diagnosis primarily relies on the results of objective auxiliary examinations, including laboratory tests, imaging examinations, and electrophysiological assessments. For instance, in the case of cardiomyopathy, the imaging patterns of echocardiography and cardiac magnetic resonance imaging (MRI) play a crucial role in the diagnostic process. This suggests a future direction that involves utilizing the large multimodal models (LMMs) to incorporate more comprehensive clinical information for diagnosis.

\subsection{Case Study}

In the evaluation of phenotype extraction using GPT-4, we observe both positive and negative cases that highlight its capabilities and limitations.

\subsubsection{Positive cases}

\begin{itemize}
    \item \textbf{High Precision in Entity Extraction:} GPT-4 adeptly avoids extracting text unrelated to the disease, such as hospital or medication details, demonstrating its discriminative capabilities.
    \item \textbf{Recognition of Negated Phenotypes:} GPT-4 accurately identifies instances where a phenotype is negated, such as correctly omitting "fever" from statements like "no fever." This indicates that GPT-4's entity extraction transcends simple keyword matching, reflecting a sophisticated comprehension of medical record texts.
\end{itemize}

\subsubsection{Negative Cases}

\begin{itemize}
    \item \textbf{Challenges in Identifying Entities with Descriptions:}  GPT-4 occasionally struggles to capture entities that include descriptive details comprehensively. For example, it reduces "Shortness of breath after walking up three flights of stairs" to merely "shortness of breath," omitting the crucial context that specifies the phenotype.
    \item \textbf{Inconsistencies in Identifying Laboratory Test Results:} GPT-4 exhibits limited accuracy in detecting abnormal laboratory test results. For instance, it failed to recognize several abnormal lab results from a patient’s medical record, including "$K\ 3.26mmol/L\downarrow$," resulting in inaccurate extraction of related phenotypes.
\end{itemize}

\section{Conclusion}

This paper delivers \textbf{\textit{RareBench}}, an innovative benchmark framework aimed at systematically assessing Large Language Models' (LLMs) capability in rare disease diagnosis. 
\textit{RareBench} combines public datasets with the comprehensive datasets from PUMCH to create the most extensive collection of rare disease patient data available. 
Leveraging an Information-Content-based random walk algorithm and a rare disease knowledge graph, we introduce a dynamic few-shot prompt, significantly enhancing LLMs' diagnostic capabilities. 
Notably, our comparative analysis with specialist physicians demonstrates GPT-4's superior accuracy in some specific rare disease diagnoses. 
We expect \textit{RareBench} to catalyze further advancements and applications of LLMs in tackling the complexities of clinical diagnosis, especially for rare diseases.

\section{Limitations and Potential Biases}
Transparency and reliability in decision-making and diagnosis are paramount in medical practice. When utilizing large language models (LLMs) in healthcare, several considerations arise, particularly regarding interpretability, reliability, and the risk of perpetuating existing biases within the models. 
Firstly, the inner workings of LLMs are often highly complex, non-transparent, and lacking interpretability, which may cause clinicians and patients to hesitate in trusting the model's decisions. Secondly, AI applications in healthcare must demonstrate high reliability and accuracy. However, due to the limitations of training data, most AI models, including LLMs, may exhibit significant prediction errors and biases, including the diagnosis of rare diseases. Additionally, AI models must justify their ability to generalize to new environments; otherwise, they may result in suboptimal performance in actual clinical settings.

LLMs may unintentionally adopt biases in their training datasets, subsequently influencing predictions. These biases are frequently derived from imbalances within medical data. This issue is particularly acute in the context of rare diseases, where data imbalances are more pronounced due to the scarcity and uneven distribution of cases. Therefore, ongoing monitoring and refinement of the model's training processes are essential to continually identify and correct biases that may emerge as the model evolves.

\section{Ethical Consideration}
Deploying LLMs in clinical settings involves several ethical considerations, including the safety, privacy, and equity of patients. Patient data is extremely sensitive. Therefore, using LLMs in such settings necessitates stringent adherence to health information privacy regulations. Although \textit{RareBench} demonstrates some promising results for LLMs in diagnosing certain rare diseases, it is essential to emphasize that LLMs should only serve as supplementary tools currently due to issues such as hallucinations. For specific diagnostic decisions, please fully heed the guidance of professional medical practitioners.

This study was approved by the Ethics Committees at Peking Union Medical College Hospital, Peking Union Medical College, and the Chinese Academy of Medical Sciences (approval number S-K2051). It is worth noting that all cases were monitored by doctors from Peking Union Medical College Hospital prior to uploading text information, ensuring the absence of any potential personal information leaks. Xuanzhong Chen, Xiaohao Mao, Qihan Guo, and Ting Chen affirm that the manuscript is an honest, accurate, and transparent account of the study being reported; that no important aspects of the study have been omitted; and that any discrepancies from the study as originally planned (and, if relevant, registered) have been explained.

\begin{acks}
This work is supported by the National Key R\&D Program of China (2021YFF1201303, 2022YFC2703103), Guoqiang Institute of Tsinghua University, and Beijing National Research Center for Information Science and Technology (BNRist). The funders had no roles in study design, data collection and analysis, the decision to publish, and manuscript preparation.
\end{acks}

\bibliographystyle{ACM-Reference-Format}
\bibliography{sample-base}


\begin{thebibliography}{63}


\ifx \showCODEN    \undefined \def \showCODEN     #1{\unskip}     \fi
\ifx \showDOI      \undefined \def \showDOI       #1{#1}\fi
\ifx \showISBNx    \undefined \def \showISBNx     #1{\unskip}     \fi
\ifx \showISBNxiii \undefined \def \showISBNxiii  #1{\unskip}     \fi
\ifx \showISSN     \undefined \def \showISSN      #1{\unskip}     \fi
\ifx \showLCCN     \undefined \def \showLCCN      #1{\unskip}     \fi
\ifx \shownote     \undefined \def \shownote      #1{#1}          \fi
\ifx \showarticletitle \undefined \def \showarticletitle #1{#1}   \fi
\ifx \showURL      \undefined \def \showURL       {\relax}        \fi
\providecommand\bibfield[2]{#2}
\providecommand\bibinfo[2]{#2}
\providecommand\natexlab[1]{#1}
\providecommand\showeprint[2][]{arXiv:#2}

\bibitem[Achiam et~al\mbox{.}(2023)]%
        {achiam2023gpt}
\bibfield{author}{\bibinfo{person}{Josh Achiam}, \bibinfo{person}{Steven Adler}, \bibinfo{person}{Sandhini Agarwal}, \bibinfo{person}{Lama Ahmad}, \bibinfo{person}{Ilge Akkaya}, \bibinfo{person}{Florencia~Leoni Aleman}, \bibinfo{person}{Diogo Almeida}, \bibinfo{person}{Janko Altenschmidt}, \bibinfo{person}{Sam Altman}, \bibinfo{person}{Shyamal Anadkat}, {et~al\mbox{.}}} \bibinfo{year}{2023}\natexlab{}.
\newblock \showarticletitle{Gpt-4 technical report}.
\newblock \bibinfo{journal}{\emph{arXiv preprint arXiv:2303.08774}} (\bibinfo{year}{2023}).
\newblock


\bibitem[Aym{\'e}(2003)]%
        {ayme2003orphanet}
\bibfield{author}{\bibinfo{person}{S{\'e}gol{\`e}ne Aym{\'e}}.} \bibinfo{year}{2003}\natexlab{}.
\newblock \showarticletitle{Orphanet, an information site on rare diseases}.
\newblock \bibinfo{journal}{\emph{Soins; la revue de r{\'e}f{\'e}rence infirmi{\`e}re}} \bibinfo{number}{672} (\bibinfo{year}{2003}), \bibinfo{pages}{46--47}.
\newblock


\bibitem[Cai et~al\mbox{.}(2023)]%
        {cai2023medbench}
\bibfield{author}{\bibinfo{person}{Yan Cai}, \bibinfo{person}{Linlin Wang}, \bibinfo{person}{Ye Wang}, \bibinfo{person}{Gerard de Melo}, \bibinfo{person}{Ya Zhang}, \bibinfo{person}{Yanfeng Wang}, {and} \bibinfo{person}{Liang He}.} \bibinfo{year}{2023}\natexlab{}.
\newblock \bibinfo{title}{MedBench: A Large-Scale Chinese Benchmark for Evaluating Medical Large Language Models}.
\newblock
\newblock
\showeprint[arxiv]{2312.12806}~[cs.CL]


\bibitem[Chen et~al\mbox{.}(2023)]%
        {chen2023huatuogptii}
\bibfield{author}{\bibinfo{person}{Junying Chen}, \bibinfo{person}{Xidong Wang}, \bibinfo{person}{Anningzhe Gao}, \bibinfo{person}{Feng Jiang}, \bibinfo{person}{Shunian Chen}, \bibinfo{person}{Hongbo Zhang}, \bibinfo{person}{Dingjie Song}, \bibinfo{person}{Wenya Xie}, \bibinfo{person}{Chuyi Kong}, \bibinfo{person}{Jianquan Li}, \bibinfo{person}{Xiang Wan}, \bibinfo{person}{Haizhou Li}, {and} \bibinfo{person}{Benyou Wang}.} \bibinfo{year}{2023}\natexlab{}.
\newblock \bibinfo{title}{HuatuoGPT-II, One-stage Training for Medical Adaption of LLMs}.
\newblock
\newblock
\showeprint[arxiv]{2311.09774}~[cs.CL]


\bibitem[Cover(1999)]%
        {cover1999elements}
\bibfield{author}{\bibinfo{person}{Thomas~M Cover}.} \bibinfo{year}{1999}\natexlab{}.
\newblock \bibinfo{booktitle}{\emph{Elements of information theory}}.
\newblock \bibinfo{publisher}{John Wiley \& Sons}.
\newblock


\bibitem[Deisseroth et~al\mbox{.}(2019)]%
        {deisseroth2019clinphen}
\bibfield{author}{\bibinfo{person}{Cole~A Deisseroth}, \bibinfo{person}{Johannes Birgmeier}, \bibinfo{person}{Ethan~E Bodle}, \bibinfo{person}{Jennefer~N Kohler}, \bibinfo{person}{Dena~R Matalon}, \bibinfo{person}{Yelena Nazarenko}, \bibinfo{person}{Casie~A Genetti}, \bibinfo{person}{Catherine~A Brownstein}, \bibinfo{person}{Klaus Schmitz-Abe}, \bibinfo{person}{Kelly Schoch}, {et~al\mbox{.}}} \bibinfo{year}{2019}\natexlab{}.
\newblock \showarticletitle{ClinPhen extracts and prioritizes patient phenotypes directly from medical records to expedite genetic disease diagnosis}.
\newblock \bibinfo{journal}{\emph{Genetics in Medicine}} \bibinfo{volume}{21}, \bibinfo{number}{7} (\bibinfo{year}{2019}), \bibinfo{pages}{1585--1593}.
\newblock


\bibitem[Du et~al\mbox{.}(2022)]%
        {du2022glm}
\bibfield{author}{\bibinfo{person}{Zhengxiao Du}, \bibinfo{person}{Yujie Qian}, \bibinfo{person}{Xiao Liu}, \bibinfo{person}{Ming Ding}, \bibinfo{person}{Jiezhong Qiu}, \bibinfo{person}{Zhilin Yang}, {and} \bibinfo{person}{Jie Tang}.} \bibinfo{year}{2022}\natexlab{}.
\newblock \showarticletitle{GLM: General Language Model Pretraining with Autoregressive Blank Infilling}. In \bibinfo{booktitle}{\emph{Proceedings of the 60th Annual Meeting of the Association for Computational Linguistics (Volume 1: Long Papers)}}. \bibinfo{pages}{320--335}.
\newblock


\bibitem[Evans(2023)]%
        {evans2023dare}
\bibfield{author}{\bibinfo{person}{William~RH Evans}.} \bibinfo{year}{2023}\natexlab{}.
\newblock \showarticletitle{Dare to think rare. Diagnostic delay and rare diseases}.
\newblock  (\bibinfo{year}{2023}).
\newblock


\bibitem[Fansi~Tchango et~al\mbox{.}(2022)]%
        {fansi2022ddxplus}
\bibfield{author}{\bibinfo{person}{Arsene Fansi~Tchango}, \bibinfo{person}{Rishab Goel}, \bibinfo{person}{Zhi Wen}, \bibinfo{person}{Julien Martel}, {and} \bibinfo{person}{Joumana Ghosn}.} \bibinfo{year}{2022}\natexlab{}.
\newblock \showarticletitle{Ddxplus: A new dataset for automatic medical diagnosis}.
\newblock \bibinfo{journal}{\emph{Advances in Neural Information Processing Systems}}  \bibinfo{volume}{35} (\bibinfo{year}{2022}), \bibinfo{pages}{31306--31318}.
\newblock


\bibitem[Feng et~al\mbox{.}(2022)]%
        {feng2022phenobert}
\bibfield{author}{\bibinfo{person}{Yuhao Feng}, \bibinfo{person}{Lei Qi}, {and} \bibinfo{person}{Weidong Tian}.} \bibinfo{year}{2022}\natexlab{}.
\newblock \showarticletitle{PhenoBERT: a combined deep learning method for automated recognition of human phenotype ontology}.
\newblock \bibinfo{journal}{\emph{IEEE/ACM Transactions on Computational Biology and Bioinformatics}} \bibinfo{volume}{20}, \bibinfo{number}{2} (\bibinfo{year}{2022}), \bibinfo{pages}{1269--1277}.
\newblock


\bibitem[Grover and Leskovec(2016)]%
        {grover2016node2vec}
\bibfield{author}{\bibinfo{person}{Aditya Grover} {and} \bibinfo{person}{Jure Leskovec}.} \bibinfo{year}{2016}\natexlab{}.
\newblock \showarticletitle{node2vec: Scalable feature learning for networks}. In \bibinfo{booktitle}{\emph{Proceedings of the 22nd ACM SIGKDD international conference on Knowledge discovery and data mining}}. \bibinfo{pages}{855--864}.
\newblock


\bibitem[Haendel et~al\mbox{.}(2020)]%
        {haendel2020many}
\bibfield{author}{\bibinfo{person}{Melissa Haendel}, \bibinfo{person}{Nicole Vasilevsky}, \bibinfo{person}{Deepak Unni}, \bibinfo{person}{Cristian Bologa}, \bibinfo{person}{Nomi Harris}, \bibinfo{person}{Heidi Rehm}, \bibinfo{person}{Ada Hamosh}, \bibinfo{person}{Gareth Baynam}, \bibinfo{person}{Tudor Groza}, \bibinfo{person}{Julie McMurry}, {et~al\mbox{.}}} \bibinfo{year}{2020}\natexlab{}.
\newblock \showarticletitle{How many rare diseases are there?}
\newblock \bibinfo{journal}{\emph{Nature reviews drug discovery}} \bibinfo{volume}{19}, \bibinfo{number}{2} (\bibinfo{year}{2020}), \bibinfo{pages}{77--78}.
\newblock


\bibitem[Hamosh et~al\mbox{.}(2005)]%
        {hamosh2005online}
\bibfield{author}{\bibinfo{person}{Ada Hamosh}, \bibinfo{person}{Alan~F Scott}, \bibinfo{person}{Joanna~S Amberger}, \bibinfo{person}{Carol~A Bocchini}, {and} \bibinfo{person}{Victor~A McKusick}.} \bibinfo{year}{2005}\natexlab{}.
\newblock \showarticletitle{Online Mendelian Inheritance in Man (OMIM), a knowledgebase of human genes and genetic disorders}.
\newblock \bibinfo{journal}{\emph{Nucleic acids research}} \bibinfo{volume}{33}, \bibinfo{number}{suppl\_1} (\bibinfo{year}{2005}), \bibinfo{pages}{D514--D517}.
\newblock


\bibitem[Han et~al\mbox{.}(2023)]%
        {han2023medalpaca}
\bibfield{author}{\bibinfo{person}{Tianyu Han}, \bibinfo{person}{Lisa~C Adams}, \bibinfo{person}{Jens-Michalis Papaioannou}, \bibinfo{person}{Paul Grundmann}, \bibinfo{person}{Tom Oberhauser}, \bibinfo{person}{Alexander L{\"o}ser}, \bibinfo{person}{Daniel Truhn}, {and} \bibinfo{person}{Keno~K Bressem}.} \bibinfo{year}{2023}\natexlab{}.
\newblock \showarticletitle{MedAlpaca--An Open-Source Collection of Medical Conversational AI Models and Training Data}.
\newblock \bibinfo{journal}{\emph{arXiv preprint arXiv:2304.08247}} (\bibinfo{year}{2023}).
\newblock


\bibitem[He et~al\mbox{.}(2019)]%
        {he2019incidence}
\bibfield{author}{\bibinfo{person}{Jiangjiang He}, \bibinfo{person}{Mi Tang}, \bibinfo{person}{Xueyan Zhang}, \bibinfo{person}{Duo Chen}, \bibinfo{person}{Qi Kang}, \bibinfo{person}{Yan Yang}, \bibinfo{person}{Jiahao Hu}, \bibinfo{person}{Chunlin Jin}, {and} \bibinfo{person}{Peipei Song}.} \bibinfo{year}{2019}\natexlab{}.
\newblock \showarticletitle{Incidence and prevalence of 121 rare diseases in China: Current status and challenges}.
\newblock \bibinfo{journal}{\emph{Intractable \& rare diseases research}} \bibinfo{volume}{8}, \bibinfo{number}{2} (\bibinfo{year}{2019}), \bibinfo{pages}{89--97}.
\newblock


\bibitem[Jia et~al\mbox{.}(2018)]%
        {jia2018rdad}
\bibfield{author}{\bibinfo{person}{Jinmeng Jia}, \bibinfo{person}{Ruiyuan Wang}, \bibinfo{person}{Zhongxin An}, \bibinfo{person}{Yongli Guo}, \bibinfo{person}{Xi Ni}, {and} \bibinfo{person}{Tieliu Shi}.} \bibinfo{year}{2018}\natexlab{}.
\newblock \showarticletitle{RDAD: a machine learning system to support phenotype-based rare disease diagnosis}.
\newblock \bibinfo{journal}{\emph{Frontiers in genetics}}  \bibinfo{volume}{9} (\bibinfo{year}{2018}), \bibinfo{pages}{587}.
\newblock


\bibitem[Jiang et~al\mbox{.}(2023)]%
        {jiang2023mistral}
\bibfield{author}{\bibinfo{person}{Albert~Q Jiang}, \bibinfo{person}{Alexandre Sablayrolles}, \bibinfo{person}{Arthur Mensch}, \bibinfo{person}{Chris Bamford}, \bibinfo{person}{Devendra~Singh Chaplot}, \bibinfo{person}{Diego de~las Casas}, \bibinfo{person}{Florian Bressand}, \bibinfo{person}{Gianna Lengyel}, \bibinfo{person}{Guillaume Lample}, \bibinfo{person}{Lucile Saulnier}, {et~al\mbox{.}}} \bibinfo{year}{2023}\natexlab{}.
\newblock \showarticletitle{Mistral 7B}.
\newblock \bibinfo{journal}{\emph{arXiv preprint arXiv:2310.06825}} (\bibinfo{year}{2023}).
\newblock


\bibitem[Jin et~al\mbox{.}(2021)]%
        {jin2021disease}
\bibfield{author}{\bibinfo{person}{Di Jin}, \bibinfo{person}{Eileen Pan}, \bibinfo{person}{Nassim Oufattole}, \bibinfo{person}{Wei-Hung Weng}, \bibinfo{person}{Hanyi Fang}, {and} \bibinfo{person}{Peter Szolovits}.} \bibinfo{year}{2021}\natexlab{}.
\newblock \showarticletitle{What disease does this patient have? a large-scale open domain question answering dataset from medical exams}.
\newblock \bibinfo{journal}{\emph{Applied Sciences}} \bibinfo{volume}{11}, \bibinfo{number}{14} (\bibinfo{year}{2021}), \bibinfo{pages}{6421}.
\newblock


\bibitem[Jin et~al\mbox{.}(2019)]%
        {jin2019pubmedqa}
\bibfield{author}{\bibinfo{person}{Qiao Jin}, \bibinfo{person}{Bhuwan Dhingra}, \bibinfo{person}{Zhengping Liu}, \bibinfo{person}{William Cohen}, {and} \bibinfo{person}{Xinghua Lu}.} \bibinfo{year}{2019}\natexlab{}.
\newblock \showarticletitle{PubMedQA: A Dataset for Biomedical Research Question Answering}. In \bibinfo{booktitle}{\emph{Proceedings of the 2019 Conference on Empirical Methods in Natural Language Processing and the 9th International Joint Conference on Natural Language Processing (EMNLP-IJCNLP)}}. \bibinfo{pages}{2567--2577}.
\newblock


\bibitem[K{\"o}hler et~al\mbox{.}(2021)]%
        {kohler2021human}
\bibfield{author}{\bibinfo{person}{Sebastian K{\"o}hler}, \bibinfo{person}{Michael Gargano}, \bibinfo{person}{Nicolas Matentzoglu}, \bibinfo{person}{Leigh~C Carmody}, \bibinfo{person}{David Lewis-Smith}, \bibinfo{person}{Nicole~A Vasilevsky}, \bibinfo{person}{Daniel Danis}, \bibinfo{person}{Ganna Balagura}, \bibinfo{person}{Gareth Baynam}, \bibinfo{person}{Amy~M Brower}, {et~al\mbox{.}}} \bibinfo{year}{2021}\natexlab{}.
\newblock \showarticletitle{The human phenotype ontology in 2021}.
\newblock \bibinfo{journal}{\emph{Nucleic acids research}} \bibinfo{volume}{49}, \bibinfo{number}{D1} (\bibinfo{year}{2021}), \bibinfo{pages}{D1207--D1217}.
\newblock


\bibitem[K{\"o}hler et~al\mbox{.}(2009)]%
        {kohler2009clinical}
\bibfield{author}{\bibinfo{person}{Sebastian K{\"o}hler}, \bibinfo{person}{Marcel~H Schulz}, \bibinfo{person}{Peter Krawitz}, \bibinfo{person}{Sebastian Bauer}, \bibinfo{person}{Sandra D{\"o}lken}, \bibinfo{person}{Claus~E Ott}, \bibinfo{person}{Christine Mundlos}, \bibinfo{person}{Denise Horn}, \bibinfo{person}{Stefan Mundlos}, {and} \bibinfo{person}{Peter~N Robinson}.} \bibinfo{year}{2009}\natexlab{}.
\newblock \showarticletitle{Clinical diagnostics in human genetics with semantic similarity searches in ontologies}.
\newblock \bibinfo{journal}{\emph{The American Journal of Human Genetics}} \bibinfo{volume}{85}, \bibinfo{number}{4} (\bibinfo{year}{2009}), \bibinfo{pages}{457--464}.
\newblock


\bibitem[K{\"o}hler et~al\mbox{.}(2017)]%
        {kohler2017human}
\bibfield{author}{\bibinfo{person}{Sebastian K{\"o}hler}, \bibinfo{person}{Nicole~A Vasilevsky}, \bibinfo{person}{Mark Engelstad}, \bibinfo{person}{Erin Foster}, \bibinfo{person}{Julie McMurry}, \bibinfo{person}{S{\'e}gol{\`e}ne Aym{\'e}}, \bibinfo{person}{Gareth Baynam}, \bibinfo{person}{Susan~M Bello}, \bibinfo{person}{Cornelius~F Boerkoel}, \bibinfo{person}{Kym~M Boycott}, {et~al\mbox{.}}} \bibinfo{year}{2017}\natexlab{}.
\newblock \showarticletitle{The human phenotype ontology in 2017}.
\newblock \bibinfo{journal}{\emph{Nucleic acids research}} \bibinfo{volume}{45}, \bibinfo{number}{D1} (\bibinfo{year}{2017}), \bibinfo{pages}{D865--D876}.
\newblock


\bibitem[Kung et~al\mbox{.}(2023)]%
        {kung2023performance}
\bibfield{author}{\bibinfo{person}{Tiffany~H Kung}, \bibinfo{person}{Morgan Cheatham}, \bibinfo{person}{Arielle Medenilla}, \bibinfo{person}{Czarina Sillos}, \bibinfo{person}{Lorie De~Leon}, \bibinfo{person}{Camille Elepa{\~n}o}, \bibinfo{person}{Maria Madriaga}, \bibinfo{person}{Rimel Aggabao}, \bibinfo{person}{Giezel Diaz-Candido}, \bibinfo{person}{James Maningo}, {et~al\mbox{.}}} \bibinfo{year}{2023}\natexlab{}.
\newblock \showarticletitle{Performance of ChatGPT on USMLE: Potential for AI-assisted medical education using large language models}.
\newblock \bibinfo{journal}{\emph{PLoS digital health}} \bibinfo{volume}{2}, \bibinfo{number}{2} (\bibinfo{year}{2023}), \bibinfo{pages}{e0000198}.
\newblock


\bibitem[Kwon et~al\mbox{.}(2023)]%
        {kwon2023large}
\bibfield{author}{\bibinfo{person}{Taeyoon Kwon}, \bibinfo{person}{Kai Tzu-iunn Ong}, \bibinfo{person}{Dongjin Kang}, \bibinfo{person}{Seungjun Moon}, \bibinfo{person}{Jeong~Ryong Lee}, \bibinfo{person}{Dosik Hwang}, \bibinfo{person}{Yongsik Sim}, \bibinfo{person}{Beomseok Sohn}, \bibinfo{person}{Dongha Lee}, {and} \bibinfo{person}{Jinyoung Yeo}.} \bibinfo{year}{2023}\natexlab{}.
\newblock \showarticletitle{Large Language Models are Clinical Reasoners: Reasoning-Aware Diagnosis Framework with Prompt-Generated Rationales}.
\newblock \bibinfo{journal}{\emph{arXiv preprint arXiv:2312.07399}} (\bibinfo{year}{2023}).
\newblock


\bibitem[Labrak et~al\mbox{.}(2024)]%
        {labrak2024biomistral}
\bibfield{author}{\bibinfo{person}{Yanis Labrak}, \bibinfo{person}{Adrien Bazoge}, \bibinfo{person}{Emmanuel Morin}, \bibinfo{person}{Pierre-Antoine Gourraud}, \bibinfo{person}{Mickael Rouvier}, {and} \bibinfo{person}{Richard Dufour}.} \bibinfo{year}{2024}\natexlab{}.
\newblock \bibinfo{title}{BioMistral: A Collection of Open-Source Pretrained Large Language Models for Medical Domains}.
\newblock
\newblock
\showeprint[arxiv]{2402.10373}~[cs.CL]


\bibitem[Lewis et~al\mbox{.}(2020)]%
        {lewis2020retrieval}
\bibfield{author}{\bibinfo{person}{Patrick Lewis}, \bibinfo{person}{Ethan Perez}, \bibinfo{person}{Aleksandra Piktus}, \bibinfo{person}{Fabio Petroni}, \bibinfo{person}{Vladimir Karpukhin}, \bibinfo{person}{Naman Goyal}, \bibinfo{person}{Heinrich K{\"u}ttler}, \bibinfo{person}{Mike Lewis}, \bibinfo{person}{Wen-tau Yih}, \bibinfo{person}{Tim Rockt{\"a}schel}, {et~al\mbox{.}}} \bibinfo{year}{2020}\natexlab{}.
\newblock \showarticletitle{Retrieval-augmented generation for knowledge-intensive nlp tasks}.
\newblock \bibinfo{journal}{\emph{Advances in Neural Information Processing Systems}}  \bibinfo{volume}{33} (\bibinfo{year}{2020}), \bibinfo{pages}{9459--9474}.
\newblock


\bibitem[Li et~al\mbox{.}(2023)]%
        {li2023ethics}
\bibfield{author}{\bibinfo{person}{Hanzhou Li}, \bibinfo{person}{John~T Moon}, \bibinfo{person}{Saptarshi Purkayastha}, \bibinfo{person}{Leo~Anthony Celi}, \bibinfo{person}{Hari Trivedi}, {and} \bibinfo{person}{Judy~W Gichoya}.} \bibinfo{year}{2023}\natexlab{}.
\newblock \showarticletitle{Ethics of large language models in medicine and medical research}.
\newblock \bibinfo{journal}{\emph{The Lancet Digital Health}} \bibinfo{volume}{5}, \bibinfo{number}{6} (\bibinfo{year}{2023}), \bibinfo{pages}{e333--e335}.
\newblock


\bibitem[Li et~al\mbox{.}(2019)]%
        {li2019xrare}
\bibfield{author}{\bibinfo{person}{Qigang Li}, \bibinfo{person}{Keyan Zhao}, \bibinfo{person}{Carlos~D Bustamante}, \bibinfo{person}{Xin Ma}, {and} \bibinfo{person}{Wing~H Wong}.} \bibinfo{year}{2019}\natexlab{}.
\newblock \showarticletitle{Xrare: a machine learning method jointly modeling phenotypes and genetic evidence for rare disease diagnosis}.
\newblock \bibinfo{journal}{\emph{Genetics in Medicine}} \bibinfo{volume}{21}, \bibinfo{number}{9} (\bibinfo{year}{2019}), \bibinfo{pages}{2126--2134}.
\newblock


\bibitem[Liu et~al\mbox{.}(2019)]%
        {liu2019doc2hpo}
\bibfield{author}{\bibinfo{person}{Cong Liu}, \bibinfo{person}{Fabricio~Sampaio Peres~Kury}, \bibinfo{person}{Ziran Li}, \bibinfo{person}{Casey Ta}, \bibinfo{person}{Kai Wang}, {and} \bibinfo{person}{Chunhua Weng}.} \bibinfo{year}{2019}\natexlab{}.
\newblock \showarticletitle{Doc2Hpo: a web application for efficient and accurate HPO concept curation}.
\newblock \bibinfo{journal}{\emph{Nucleic acids research}} \bibinfo{volume}{47}, \bibinfo{number}{W1} (\bibinfo{year}{2019}), \bibinfo{pages}{W566--W570}.
\newblock


\bibitem[Liu et~al\mbox{.}(2023b)]%
        {liu2023benchmarking}
\bibfield{author}{\bibinfo{person}{Junling Liu}, \bibinfo{person}{Peilin Zhou}, \bibinfo{person}{Yining Hua}, \bibinfo{person}{Dading Chong}, \bibinfo{person}{Zhongyu Tian}, \bibinfo{person}{Andrew Liu}, \bibinfo{person}{Helin Wang}, \bibinfo{person}{Chenyu You}, \bibinfo{person}{Zhenhua Guo}, \bibinfo{person}{Lei Zhu}, {et~al\mbox{.}}} \bibinfo{year}{2023}\natexlab{b}.
\newblock \showarticletitle{Benchmarking Large Language Models on CMExam--A Comprehensive Chinese Medical Exam Dataset}.
\newblock \bibinfo{journal}{\emph{arXiv preprint arXiv:2306.03030}} (\bibinfo{year}{2023}).
\newblock


\bibitem[Liu et~al\mbox{.}(2023a)]%
        {liu2023exploring}
\bibfield{author}{\bibinfo{person}{Qianchu Liu}, \bibinfo{person}{Stephanie Hyland}, \bibinfo{person}{Shruthi Bannur}, \bibinfo{person}{Kenza Bouzid}, \bibinfo{person}{Daniel Castro}, \bibinfo{person}{Maria Wetscherek}, \bibinfo{person}{Robert Tinn}, \bibinfo{person}{Harshita Sharma}, \bibinfo{person}{Fernando P{\'e}rez-Garc{\'\i}a}, \bibinfo{person}{Anton Schwaighofer}, {et~al\mbox{.}}} \bibinfo{year}{2023}\natexlab{a}.
\newblock \showarticletitle{Exploring the Boundaries of GPT-4 in Radiology}. In \bibinfo{booktitle}{\emph{Proceedings of the 2023 Conference on Empirical Methods in Natural Language Processing}}. \bibinfo{pages}{14414--14445}.
\newblock


\bibitem[Luo et~al\mbox{.}(2021)]%
        {luo2021phenotagger}
\bibfield{author}{\bibinfo{person}{Ling Luo}, \bibinfo{person}{Shankai Yan}, \bibinfo{person}{Po-Ting Lai}, \bibinfo{person}{Daniel Veltri}, \bibinfo{person}{Andrew Oler}, \bibinfo{person}{Sandhya Xirasagar}, \bibinfo{person}{Rajarshi Ghosh}, \bibinfo{person}{Morgan Similuk}, \bibinfo{person}{Peter~N Robinson}, {and} \bibinfo{person}{Zhiyong Lu}.} \bibinfo{year}{2021}\natexlab{}.
\newblock \showarticletitle{PhenoTagger: a hybrid method for phenotype concept recognition using human phenotype ontology}.
\newblock \bibinfo{journal}{\emph{Bioinformatics}} \bibinfo{volume}{37}, \bibinfo{number}{13} (\bibinfo{year}{2021}), \bibinfo{pages}{1884--1890}.
\newblock


\bibitem[Marwaha et~al\mbox{.}(2022)]%
        {marwaha2022guide}
\bibfield{author}{\bibinfo{person}{Shruti Marwaha}, \bibinfo{person}{Joshua~W Knowles}, {and} \bibinfo{person}{Euan~A Ashley}.} \bibinfo{year}{2022}\natexlab{}.
\newblock \showarticletitle{A guide for the diagnosis of rare and undiagnosed disease: beyond the exome}.
\newblock \bibinfo{journal}{\emph{Genome medicine}} \bibinfo{volume}{14}, \bibinfo{number}{1} (\bibinfo{year}{2022}), \bibinfo{pages}{1--22}.
\newblock


\bibitem[McDuff et~al\mbox{.}(2023)]%
        {mcduff2023towards}
\bibfield{author}{\bibinfo{person}{Daniel McDuff}, \bibinfo{person}{Mike Schaekermann}, \bibinfo{person}{Tao Tu}, \bibinfo{person}{Anil Palepu}, \bibinfo{person}{Amy Wang}, \bibinfo{person}{Jake Garrison}, \bibinfo{person}{Karan Singhal}, \bibinfo{person}{Yash Sharma}, \bibinfo{person}{Shekoofeh Azizi}, \bibinfo{person}{Kavita Kulkarni}, {et~al\mbox{.}}} \bibinfo{year}{2023}\natexlab{}.
\newblock \showarticletitle{Towards accurate differential diagnosis with large language models}.
\newblock \bibinfo{journal}{\emph{arXiv preprint arXiv:2312.00164}} (\bibinfo{year}{2023}).
\newblock


\bibitem[Mikolov et~al\mbox{.}(2013)]%
        {mikolov2013distributed}
\bibfield{author}{\bibinfo{person}{Tomas Mikolov}, \bibinfo{person}{Ilya Sutskever}, \bibinfo{person}{Kai Chen}, \bibinfo{person}{Greg~S Corrado}, {and} \bibinfo{person}{Jeff Dean}.} \bibinfo{year}{2013}\natexlab{}.
\newblock \showarticletitle{Distributed representations of words and phrases and their compositionality}.
\newblock \bibinfo{journal}{\emph{Advances in neural information processing systems}}  \bibinfo{volume}{26} (\bibinfo{year}{2013}).
\newblock


\bibitem[Moor et~al\mbox{.}(2023)]%
        {moor2023foundation}
\bibfield{author}{\bibinfo{person}{Michael Moor}, \bibinfo{person}{Oishi Banerjee}, \bibinfo{person}{Zahra Shakeri~Hossein Abad}, \bibinfo{person}{Harlan~M Krumholz}, \bibinfo{person}{Jure Leskovec}, \bibinfo{person}{Eric~J Topol}, {and} \bibinfo{person}{Pranav Rajpurkar}.} \bibinfo{year}{2023}\natexlab{}.
\newblock \showarticletitle{Foundation models for generalist medical artificial intelligence}.
\newblock \bibinfo{journal}{\emph{Nature}} \bibinfo{volume}{616}, \bibinfo{number}{7956} (\bibinfo{year}{2023}), \bibinfo{pages}{259--265}.
\newblock


\bibitem[Nori et~al\mbox{.}(2023a)]%
        {nori2023capabilities}
\bibfield{author}{\bibinfo{person}{Harsha Nori}, \bibinfo{person}{Nicholas King}, \bibinfo{person}{Scott~Mayer McKinney}, \bibinfo{person}{Dean Carignan}, {and} \bibinfo{person}{Eric Horvitz}.} \bibinfo{year}{2023}\natexlab{a}.
\newblock \showarticletitle{Capabilities of gpt-4 on medical challenge problems}.
\newblock \bibinfo{journal}{\emph{arXiv preprint arXiv:2303.13375}} (\bibinfo{year}{2023}).
\newblock


\bibitem[Nori et~al\mbox{.}(2023b)]%
        {nori2023can}
\bibfield{author}{\bibinfo{person}{Harsha Nori}, \bibinfo{person}{Yin~Tat Lee}, \bibinfo{person}{Sheng Zhang}, \bibinfo{person}{Dean Carignan}, \bibinfo{person}{Richard Edgar}, \bibinfo{person}{Nicolo Fusi}, \bibinfo{person}{Nicholas King}, \bibinfo{person}{Jonathan Larson}, \bibinfo{person}{Yuanzhi Li}, \bibinfo{person}{Weishung Liu}, {et~al\mbox{.}}} \bibinfo{year}{2023}\natexlab{b}.
\newblock \showarticletitle{Can generalist foundation models outcompete special-purpose tuning? case study in medicine}.
\newblock \bibinfo{journal}{\emph{arXiv preprint arXiv:2311.16452}} (\bibinfo{year}{2023}).
\newblock


\bibitem[OpenAI(2022a)]%
        {noauthor_introducing_nodate}
\bibfield{author}{\bibinfo{person}{OpenAI}.} \bibinfo{year}{2022}\natexlab{a}.
\newblock \bibinfo{title}{Introducing {ChatGPT}}.
\newblock
\newblock
\urldef\tempurl%
\url{https://openai.com/blog/chatgpt}
\showURL{%
\tempurl}


\bibitem[OpenAI(2022b)]%
        {openai_embedding}
\bibfield{author}{\bibinfo{person}{OpenAI}.} \bibinfo{year}{2022}\natexlab{b}.
\newblock \bibinfo{title}{New and improved embedding model}.
\newblock
\newblock
\urldef\tempurl%
\url{https://openai.com/blog/new-and-improved-embedding-model}
\showURL{%
\tempurl}


\bibitem[Pal et~al\mbox{.}(2022)]%
        {pal2022medmcqa}
\bibfield{author}{\bibinfo{person}{Ankit Pal}, \bibinfo{person}{Logesh~Kumar Umapathi}, {and} \bibinfo{person}{Malaikannan Sankarasubbu}.} \bibinfo{year}{2022}\natexlab{}.
\newblock \showarticletitle{Medmcqa: A large-scale multi-subject multi-choice dataset for medical domain question answering}. In \bibinfo{booktitle}{\emph{Conference on Health, Inference, and Learning}}. PMLR, \bibinfo{pages}{248--260}.
\newblock


\bibitem[Peng et~al\mbox{.}(2016)]%
        {peng2016measuring}
\bibfield{author}{\bibinfo{person}{Jiajie Peng}, \bibinfo{person}{Hansheng Xue}, \bibinfo{person}{Yukai Shao}, \bibinfo{person}{Xuequn Shang}, \bibinfo{person}{Yadong Wang}, {and} \bibinfo{person}{Jin Chen}.} \bibinfo{year}{2016}\natexlab{}.
\newblock \showarticletitle{Measuring phenotype semantic similarity using human phenotype ontology}. In \bibinfo{booktitle}{\emph{2016 IEEE International Conference on Bioinformatics and Biomedicine (BIBM)}}. IEEE, \bibinfo{pages}{763--766}.
\newblock


\bibitem[Perozzi et~al\mbox{.}(2014)]%
        {perozzi2014deepwalk}
\bibfield{author}{\bibinfo{person}{Bryan Perozzi}, \bibinfo{person}{Rami Al-Rfou}, {and} \bibinfo{person}{Steven Skiena}.} \bibinfo{year}{2014}\natexlab{}.
\newblock \showarticletitle{Deepwalk: Online learning of social representations}. In \bibinfo{booktitle}{\emph{Proceedings of the 20th ACM SIGKDD international conference on Knowledge discovery and data mining}}. \bibinfo{pages}{701--710}.
\newblock


\bibitem[Philippakis et~al\mbox{.}(2015)]%
        {philippakis2015matchmaker}
\bibfield{author}{\bibinfo{person}{Anthony~A Philippakis}, \bibinfo{person}{Danielle~R Azzariti}, \bibinfo{person}{Sergi Beltran}, \bibinfo{person}{Anthony~J Brookes}, \bibinfo{person}{Catherine~A Brownstein}, \bibinfo{person}{Michael Brudno}, \bibinfo{person}{Han~G Brunner}, \bibinfo{person}{Orion~J Buske}, \bibinfo{person}{Knox Carey}, \bibinfo{person}{Cassie Doll}, {et~al\mbox{.}}} \bibinfo{year}{2015}\natexlab{}.
\newblock \showarticletitle{The Matchmaker Exchange: a platform for rare disease gene discovery}.
\newblock \bibinfo{journal}{\emph{Human mutation}} \bibinfo{volume}{36}, \bibinfo{number}{10} (\bibinfo{year}{2015}), \bibinfo{pages}{915--921}.
\newblock


\bibitem[Pinol et~al\mbox{.}(2017)]%
        {pinol2017rare}
\bibfield{author}{\bibinfo{person}{Marc Pinol}, \bibinfo{person}{Rui Alves}, \bibinfo{person}{Ivan Teixido}, \bibinfo{person}{Jordi Mateo}, \bibinfo{person}{Francesc Solsona}, {and} \bibinfo{person}{Ester Vilapriny{\'o}}.} \bibinfo{year}{2017}\natexlab{}.
\newblock \showarticletitle{Rare disease discovery: An optimized disease ranking system}.
\newblock \bibinfo{journal}{\emph{IEEE Transactions on Industrial Informatics}} \bibinfo{volume}{13}, \bibinfo{number}{3} (\bibinfo{year}{2017}), \bibinfo{pages}{1184--1192}.
\newblock


\bibitem[Robinson et~al\mbox{.}(2020)]%
        {robinson2020interpretable}
\bibfield{author}{\bibinfo{person}{Peter~N Robinson}, \bibinfo{person}{Vida Ravanmehr}, \bibinfo{person}{Julius~OB Jacobsen}, \bibinfo{person}{Daniel Danis}, \bibinfo{person}{Xingmin~Aaron Zhang}, \bibinfo{person}{Leigh~C Carmody}, \bibinfo{person}{Michael~A Gargano}, \bibinfo{person}{Courtney~L Thaxton}, \bibinfo{person}{Guy Karlebach}, \bibinfo{person}{Justin Reese}, {et~al\mbox{.}}} \bibinfo{year}{2020}\natexlab{}.
\newblock \showarticletitle{Interpretable clinical genomics with a likelihood ratio paradigm}.
\newblock \bibinfo{journal}{\emph{The American Journal of Human Genetics}} \bibinfo{volume}{107}, \bibinfo{number}{3} (\bibinfo{year}{2020}), \bibinfo{pages}{403--417}.
\newblock


\bibitem[Ronicke et~al\mbox{.}(2019)]%
        {ronicke2019can}
\bibfield{author}{\bibinfo{person}{Simon Ronicke}, \bibinfo{person}{Martin~C Hirsch}, \bibinfo{person}{Ewelina T{\"u}rk}, \bibinfo{person}{Katharina Larionov}, \bibinfo{person}{Daphne Tientcheu}, {and} \bibinfo{person}{Annette~D Wagner}.} \bibinfo{year}{2019}\natexlab{}.
\newblock \showarticletitle{Can a decision support system accelerate rare disease diagnosis? Evaluating the potential impact of Ada DX in a retrospective study}.
\newblock \bibinfo{journal}{\emph{Orphanet journal of rare diseases}}  \bibinfo{volume}{14} (\bibinfo{year}{2019}), \bibinfo{pages}{1--12}.
\newblock


\bibitem[Singhal et~al\mbox{.}(2023)]%
        {singhal2023large}
\bibfield{author}{\bibinfo{person}{Karan Singhal}, \bibinfo{person}{Shekoofeh Azizi}, \bibinfo{person}{Tao Tu}, \bibinfo{person}{S~Sara Mahdavi}, \bibinfo{person}{Jason Wei}, \bibinfo{person}{Hyung~Won Chung}, \bibinfo{person}{Nathan Scales}, \bibinfo{person}{Ajay Tanwani}, \bibinfo{person}{Heather Cole-Lewis}, \bibinfo{person}{Stephen Pfohl}, {et~al\mbox{.}}} \bibinfo{year}{2023}\natexlab{}.
\newblock \showarticletitle{Large language models encode clinical knowledge}.
\newblock \bibinfo{journal}{\emph{Nature}} \bibinfo{volume}{620}, \bibinfo{number}{7972} (\bibinfo{year}{2023}), \bibinfo{pages}{172--180}.
\newblock


\bibitem[Son et~al\mbox{.}(2018)]%
        {son2018deep}
\bibfield{author}{\bibinfo{person}{Jung~Hoon Son}, \bibinfo{person}{Gangcai Xie}, \bibinfo{person}{Chi Yuan}, \bibinfo{person}{Lyudmila Ena}, \bibinfo{person}{Ziran Li}, \bibinfo{person}{Andrew Goldstein}, \bibinfo{person}{Lulin Huang}, \bibinfo{person}{Liwei Wang}, \bibinfo{person}{Feichen Shen}, \bibinfo{person}{Hongfang Liu}, {et~al\mbox{.}}} \bibinfo{year}{2018}\natexlab{}.
\newblock \showarticletitle{Deep phenotyping on electronic health records facilitates genetic diagnosis by clinical exomes}.
\newblock \bibinfo{journal}{\emph{The American Journal of Human Genetics}} \bibinfo{volume}{103}, \bibinfo{number}{1} (\bibinfo{year}{2018}), \bibinfo{pages}{58--73}.
\newblock


\bibitem[Team et~al\mbox{.}(2023)]%
        {team2023gemini}
\bibfield{author}{\bibinfo{person}{Gemini Team}, \bibinfo{person}{Rohan Anil}, \bibinfo{person}{Sebastian Borgeaud}, \bibinfo{person}{Yonghui Wu}, \bibinfo{person}{Jean-Baptiste Alayrac}, \bibinfo{person}{Jiahui Yu}, \bibinfo{person}{Radu Soricut}, \bibinfo{person}{Johan Schalkwyk}, \bibinfo{person}{Andrew~M Dai}, \bibinfo{person}{Anja Hauth}, {et~al\mbox{.}}} \bibinfo{year}{2023}\natexlab{}.
\newblock \showarticletitle{Gemini: a family of highly capable multimodal models}.
\newblock \bibinfo{journal}{\emph{arXiv preprint arXiv:2312.11805}} (\bibinfo{year}{2023}).
\newblock


\bibitem[Thirunavukarasu et~al\mbox{.}(2023)]%
        {thirunavukarasu2023large}
\bibfield{author}{\bibinfo{person}{Arun~James Thirunavukarasu}, \bibinfo{person}{Darren Shu~Jeng Ting}, \bibinfo{person}{Kabilan Elangovan}, \bibinfo{person}{Laura Gutierrez}, \bibinfo{person}{Ting~Fang Tan}, {and} \bibinfo{person}{Daniel Shu~Wei Ting}.} \bibinfo{year}{2023}\natexlab{}.
\newblock \showarticletitle{Large language models in medicine}.
\newblock \bibinfo{journal}{\emph{Nature medicine}} \bibinfo{volume}{29}, \bibinfo{number}{8} (\bibinfo{year}{2023}), \bibinfo{pages}{1930--1940}.
\newblock


\bibitem[T{\"o}pel et~al\mbox{.}(2010)]%
        {topel2010ramedis}
\bibfield{author}{\bibinfo{person}{Thoralf T{\"o}pel}, \bibinfo{person}{Dagmar Scheible}, \bibinfo{person}{Friedrich Trefz}, {and} \bibinfo{person}{Ralf Hofest{\"a}dt}.} \bibinfo{year}{2010}\natexlab{}.
\newblock \showarticletitle{RAMEDIS: a comprehensive information system for variations and corresponding phenotypes of rare metabolic diseases}.
\newblock \bibinfo{journal}{\emph{Human mutation}} \bibinfo{volume}{31}, \bibinfo{number}{1} (\bibinfo{year}{2010}), \bibinfo{pages}{E1081--E1088}.
\newblock


\bibitem[Touvron et~al\mbox{.}(2023)]%
        {touvron2023llama}
\bibfield{author}{\bibinfo{person}{Hugo Touvron}, \bibinfo{person}{Louis Martin}, \bibinfo{person}{Kevin Stone}, \bibinfo{person}{Peter Albert}, \bibinfo{person}{Amjad Almahairi}, \bibinfo{person}{Yasmine Babaei}, \bibinfo{person}{Nikolay Bashlykov}, \bibinfo{person}{Soumya Batra}, \bibinfo{person}{Prajjwal Bhargava}, \bibinfo{person}{Shruti Bhosale}, {et~al\mbox{.}}} \bibinfo{year}{2023}\natexlab{}.
\newblock \showarticletitle{Llama 2: Open foundation and fine-tuned chat models}.
\newblock \bibinfo{journal}{\emph{arXiv preprint arXiv:2307.09288}} (\bibinfo{year}{2023}).
\newblock


\bibitem[Tu et~al\mbox{.}(2024)]%
        {tu2024towards}
\bibfield{author}{\bibinfo{person}{Tao Tu}, \bibinfo{person}{Anil Palepu}, \bibinfo{person}{Mike Schaekermann}, \bibinfo{person}{Khaled Saab}, \bibinfo{person}{Jan Freyberg}, \bibinfo{person}{Ryutaro Tanno}, \bibinfo{person}{Amy Wang}, \bibinfo{person}{Brenna Li}, \bibinfo{person}{Mohamed Amin}, \bibinfo{person}{Nenad Tomasev}, {et~al\mbox{.}}} \bibinfo{year}{2024}\natexlab{}.
\newblock \showarticletitle{Towards Conversational Diagnostic AI}.
\newblock \bibinfo{journal}{\emph{arXiv preprint arXiv:2401.05654}} (\bibinfo{year}{2024}).
\newblock


\bibitem[Vaid et~al\mbox{.}(2023)]%
        {vaid2023using}
\bibfield{author}{\bibinfo{person}{Akhil Vaid}, \bibinfo{person}{Isotta Landi}, \bibinfo{person}{Girish Nadkarni}, {and} \bibinfo{person}{Ismail Nabeel}.} \bibinfo{year}{2023}\natexlab{}.
\newblock \showarticletitle{Using fine-tuned large language models to parse clinical notes in musculoskeletal pain disorders}.
\newblock \bibinfo{journal}{\emph{The Lancet Digital Health}} \bibinfo{volume}{5}, \bibinfo{number}{12} (\bibinfo{year}{2023}), \bibinfo{pages}{e855--e858}.
\newblock


\bibitem[Wang et~al\mbox{.}(2023)]%
        {wang2023chatcad}
\bibfield{author}{\bibinfo{person}{Sheng Wang}, \bibinfo{person}{Zihao Zhao}, \bibinfo{person}{Xi Ouyang}, \bibinfo{person}{Qian Wang}, {and} \bibinfo{person}{Dinggang Shen}.} \bibinfo{year}{2023}\natexlab{}.
\newblock \showarticletitle{Chatcad: Interactive computer-aided diagnosis on medical image using large language models}.
\newblock \bibinfo{journal}{\emph{arXiv preprint arXiv:2302.07257}} (\bibinfo{year}{2023}).
\newblock


\bibitem[Wei et~al\mbox{.}(2022)]%
        {wei2022chain}
\bibfield{author}{\bibinfo{person}{Jason Wei}, \bibinfo{person}{Xuezhi Wang}, \bibinfo{person}{Dale Schuurmans}, \bibinfo{person}{Maarten Bosma}, \bibinfo{person}{Fei Xia}, \bibinfo{person}{Ed Chi}, \bibinfo{person}{Quoc~V Le}, \bibinfo{person}{Denny Zhou}, {et~al\mbox{.}}} \bibinfo{year}{2022}\natexlab{}.
\newblock \showarticletitle{Chain-of-thought prompting elicits reasoning in large language models}.
\newblock \bibinfo{journal}{\emph{Advances in Neural Information Processing Systems}}  \bibinfo{volume}{35} (\bibinfo{year}{2022}), \bibinfo{pages}{24824--24837}.
\newblock


\bibitem[Yang et~al\mbox{.}(2023)]%
        {yang2023enhancing}
\bibfield{author}{\bibinfo{person}{Jingye Yang}, \bibinfo{person}{Cong Liu}, \bibinfo{person}{Wendy Deng}, \bibinfo{person}{Da Wu}, \bibinfo{person}{Chunhua Weng}, \bibinfo{person}{Yunyun Zhou}, {and} \bibinfo{person}{Kai Wang}.} \bibinfo{year}{2023}\natexlab{}.
\newblock \showarticletitle{Enhancing phenotype recognition in clinical notes using large language models: PhenoBCBERT and PhenoGPT}.
\newblock \bibinfo{journal}{\emph{Patterns}} (\bibinfo{year}{2023}).
\newblock


\bibitem[Yang et~al\mbox{.}(2022)]%
        {yang2022large}
\bibfield{author}{\bibinfo{person}{Xi Yang}, \bibinfo{person}{Aokun Chen}, \bibinfo{person}{Nima PourNejatian}, \bibinfo{person}{Hoo~Chang Shin}, \bibinfo{person}{Kaleb~E Smith}, \bibinfo{person}{Christopher Parisien}, \bibinfo{person}{Colin Compas}, \bibinfo{person}{Cheryl Martin}, \bibinfo{person}{Anthony~B Costa}, \bibinfo{person}{Mona~G Flores}, {et~al\mbox{.}}} \bibinfo{year}{2022}\natexlab{}.
\newblock \showarticletitle{A large language model for electronic health records}.
\newblock \bibinfo{journal}{\emph{NPJ Digital Medicine}} \bibinfo{volume}{5}, \bibinfo{number}{1} (\bibinfo{year}{2022}), \bibinfo{pages}{194}.
\newblock


\bibitem[Zeng et~al\mbox{.}(2022)]%
        {zeng2022glm}
\bibfield{author}{\bibinfo{person}{Aohan Zeng}, \bibinfo{person}{Xiao Liu}, \bibinfo{person}{Zhengxiao Du}, \bibinfo{person}{Zihan Wang}, \bibinfo{person}{Hanyu Lai}, \bibinfo{person}{Ming Ding}, \bibinfo{person}{Zhuoyi Yang}, \bibinfo{person}{Yifan Xu}, \bibinfo{person}{Wendi Zheng}, \bibinfo{person}{Xiao Xia}, {et~al\mbox{.}}} \bibinfo{year}{2022}\natexlab{}.
\newblock \showarticletitle{Glm-130b: An open bilingual pre-trained model}.
\newblock \bibinfo{journal}{\emph{arXiv preprint arXiv:2210.02414}} (\bibinfo{year}{2022}).
\newblock


\bibitem[Zhai et~al\mbox{.}(2023)]%
        {zhai2023phen2disease}
\bibfield{author}{\bibinfo{person}{Weiqi Zhai}, \bibinfo{person}{Xiaodi Huang}, \bibinfo{person}{Nan Shen}, {and} \bibinfo{person}{Shanfeng Zhu}.} \bibinfo{year}{2023}\natexlab{}.
\newblock \showarticletitle{Phen2Disease: a phenotype-driven model for disease and gene prioritization by bidirectional maximum matching semantic similarities}.
\newblock \bibinfo{journal}{\emph{Briefings in Bioinformatics}} (\bibinfo{year}{2023}), \bibinfo{pages}{bbad172}.
\newblock


\bibitem[Zhang et~al\mbox{.}(2022)]%
        {zhang2022automatic}
\bibfield{author}{\bibinfo{person}{Zhuosheng Zhang}, \bibinfo{person}{Aston Zhang}, \bibinfo{person}{Mu Li}, {and} \bibinfo{person}{Alex Smola}.} \bibinfo{year}{2022}\natexlab{}.
\newblock \showarticletitle{Automatic chain of thought prompting in large language models}.
\newblock \bibinfo{journal}{\emph{arXiv preprint arXiv:2210.03493}} (\bibinfo{year}{2022}).
\newblock


\bibitem[Zhao et~al\mbox{.}(2020)]%
        {zhao2020phen2gene}
\bibfield{author}{\bibinfo{person}{Mengge Zhao}, \bibinfo{person}{James~M Havrilla}, \bibinfo{person}{Li Fang}, \bibinfo{person}{Ying Chen}, \bibinfo{person}{Jacqueline Peng}, \bibinfo{person}{Cong Liu}, \bibinfo{person}{Chao Wu}, \bibinfo{person}{Mahdi Sarmady}, \bibinfo{person}{Pablo Botas}, \bibinfo{person}{Juli{\'a}n Isla}, {et~al\mbox{.}}} \bibinfo{year}{2020}\natexlab{}.
\newblock \showarticletitle{Phen2Gene: rapid phenotype-driven gene prioritization for rare diseases}.
\newblock \bibinfo{journal}{\emph{NAR genomics and Bioinformatics}} \bibinfo{volume}{2}, \bibinfo{number}{2} (\bibinfo{year}{2020}), \bibinfo{pages}{lqaa032}.
\newblock


\end{thebibliography}

\appendix

\definecolor{examplegray}{gray}{0.9} 

\section{Appendix}
\label{adp}
\subsection{Dataset Details}
\label{app:dataset}
Four publicly available datasets are used in this study: MME\cite{philippakis2015matchmaker}, LIRICAL\cite{robinson2020interpretable}, HMS\cite{ronicke2019can}, and RAMEDIS\cite{topel2010ramedis}. These datasets are sourced from published articles or open-access datasets. Quality control measures were implemented to filter out cases with insufficient information. Following screening procedures, the MME\footnote {https://github.com/ga4gh/mme-apis/tree/master/testing} dataset comprises 40 cases across 17 diseases, the LIRICAL\footnote{https://
zenodo.org/record/3905420} dataset encompasses 370 cases spanning 252 diseases, the HMS dataset encompasses 88 cases covering 39 diseases, and the RAMEDIS dataset consists of 624 cases spanning 63 diseases.

\subsection{Framework Settings}
\label{app:framework}

The implementation details of the models listed in Table \ref{tab:model_description} are as follows: 
Both GPT-4 and GPT-3.5 are accessed through OpenAI's API \footnote{https://platform.openai.com/docs/introduction}. The models used are "gpt-4-1106-preview" and "gpt-3.5-turbo-1106" respectively. For the two models, the seed parameter is set to 42, while all other parameters are left at their default settings. 
Both GLM4 and GLM3 are accessed through Zhipu AI's API \footnote{https://open.bigmodel.cn/}. The models used are "glm-4" and "glm-3-turbo" respectively. In the model parameter settings, temperature is set to 0.15 and top\_p to 0.7; all other parameters are maintained at their default values. 
Gemini is accessed through DeepMind's API \footnote{https://deepmind.google/technologies/gemini/\#introduction}. The model used is "gemini-pro", and the parameters use default settings.

\subsection{Prompt Examples}
\subsubsection{Task 1: Phenotype Extraction from Electronic Health Records}
~

Sub-task 1: \textit{Phenotype Extraction}

\begin{framed}
\scriptsize
\noindent\colorbox{examplegray}{
[Task]
}
Extract disease phenotypes from the following medical record text.

\noindent\colorbox{examplegray}{
[Requirement]
}
\begin{enumerate}[leftmargin=*]
    \item Output format: One phenotype per line, the format is [index]. [English name], such as "1. Cough", "2. Pain".
    \item  Answer according to the standard English terminology in the HPO database (https://hpo.jax.org/app/). Do not use colloquialisms, and try to be as concise as possible. It is not a restatement of the original words, but a refinement of the phenotype.
    \item Extract all phenotypes appearing in the text without omitting any. output as much as possible.
    \item For symptoms that are denied in the text, such as "no chest pain," and "no cough," do not extract the corresponding phenotypes.
\end{enumerate}

\noindent\colorbox{examplegray}{
[Medical Record Text]
}

The patient developed shortness of breath accompanied by chest tightness and facial edema about 3 months ago after physical exertion, with the eyelid edema being the most severe. The shortness of breath was most severe after physical activity and improved after rest. He was treated at a local hospital, and on December 1, 2013, the cardiac function showed by color echocardiography: tricuspid valve disease. Then he went to our hospital. Ultrasound showed myocardial disease, right atrium and right ventricle enlargement, moderate to severe tricuspid valve insufficiency, severe reduction in left and right ventricular systolic function, aortic valve degeneration, and a small amount of pericardial effusion. The patient was admitted to our hospital for outpatient consideration of tricuspid valve disease and is now admitted to our department for further diagnosis and treatment. Since the onset of the disease, the patient's energy, diet, and sleep have been acceptable, and his bowel movements have been normal. There is no significant change in weight from before. No dry mouth, dry eyes, mouth ulcers, joint swelling and pain, rash, etc.

\end{framed}

Sub-task 2: \textit{General Entity Extraction}

\begin{framed}
\scriptsize
\noindent\colorbox{examplegray}{
[Task]
}
For the medical record text provided below, mark the text that represents the disease entity.

\noindent\colorbox{examplegray}{
[Requirement]
}

\begin{enumerate}[leftmargin=*]
    \item  Output format: One disease entity per line. The format is [Index]. [Original text]. For example, "1. Fever", "2. Body temperature 39℃".
    \item Output in the order in which it appears in the text; do not omit anything, and output as much as possible.
    \item For symptoms that are denied in the text, such as "no chest pain," and "no cough," do not extract the corresponding entities.
\end{enumerate}

\noindent\colorbox{examplegray}{
[Medical Record Text]
}

$\cdots$

\end{framed}

Sub-task3: \textit{General Entity Standardization}

\begin{framed}
\scriptsize
\noindent\colorbox{examplegray}{
[Task]
}
Standardize the following medical entities to Human Phenotype Ontology phenotypes.

\noindent\colorbox{examplegray}{
[Requirement]
}

\begin{enumerate}[leftmargin=*]
    \item Answer according to the standard English terminology in the HPO database (https://hpo.jax.org/app/). Do not be colloquial, and try to be as formal, standardized, and concise as possible.
    \item Output format: Each line contains one medical entity and its corresponding HPO phenotype. The format is [index]. [Entity name]: [English name of phenotype]. For example, "1. Body temperature 39℃: Fever", "2. Excessive urine output: Polyuria".
\end{enumerate}

\noindent\colorbox{examplegray}{
[Entity list]
}

Chest tightness

ST segment elevation

Activity tolerance gradually decreases

ventricular fibrillation

Aspen syndrome

sore throat

J-point elevation greater than 0.2mv and saddle-like elevation

Shortness of breath

pneumonia

cough
\end{framed}

\subsubsection{Task 2: Screening for Specific Rare Diseases}
~

\begin{framed}
\scriptsize
\noindent\colorbox{examplegray}{
Zero-shot prompt
}
As an expert in the field of rare diseases, you are tasked with diagnosing a real clinical case. Please carefully review the patient's basic medical history, specialized examinations, and auxiliary tests to determine whether the patient may be suffering from [SCREENING DISEASE].

\noindent\colorbox{examplegray}{
CoT prompt
}
[Zero-shot prompt], Let us think step by step.
\end{framed}

\subsubsection{Task 3: Comparison Analysis of Common and Rare Diseases}
~

\begin{framed}
\scriptsize
\noindent\colorbox{examplegray}{
Zero-shot prompt
}
Please, as a rare disease specialist, answer the following questions. [EHRs] is the patient's admitted record, including chief complaint, present medical history, etc. [77 CANDIDATE DISEASES] are the types of diseases that the patient may suffer from. The top 10 diagnosed diseases are selected from highest to lowest probability.

\noindent\colorbox{examplegray}{
CoT prompt
}
[Zero-shot prompt], Let us think step by step.

\end{framed}

\subsubsection{Task 4: Differential Diagnosis among Universal Rare Diseases}

Here, we present the specific configurations for zero-shot, Chain of Thought (CoT), and (dynamic) few-shot settings.

\begin{framed}
\scriptsize
\noindent\colorbox{examplegray}{
System prompt
}
You are a specialist in the field of rare diseases. You will be provided and asked about a complicated clinical case; read it carefully and then provide a diverse and comprehensive differential diagnosis.

\noindent\colorbox{examplegray}{
Zero-shot prompt
}
This rare disease patient suffers from symptoms: [PATIENT\_PHENOTYPE].
Enumerate the top 10 most likely diagnoses. Be precise, listing one diagnosis per line, and try to cover many unique possibilities (at least 10). The top 10 diagnoses are:

\noindent\colorbox{examplegray}{
CoT prompt
}
[Zero-shot prompt].
Let us think step by step.

\noindent\colorbox{examplegray}{
(Dynamic) Few-shot prompt
}
Let me give you [K] examples first: The [i] patient has a rare disease [EXAMPLE\_DIAGNOSIS], and his/her [PHENOTYPE or EHR] is as follows: [EXAMPLE\_PHENOTYPE or EXAMPLE\_EHR]. Next is the patient case you need to diagnose:  [Zero-shot prompt].

\end{framed}

\subsection{Knowledge Integration Dynamic Few-shot}
\label{app:knowledge}

\subsubsection{Knowledge Bases Details}

The Human Phenotype Ontology (HPO) \footnote{https://hpo.jax.org/app/} provides a standardized vocabulary for phenotypic abnormalities encountered in human disease. 
These phenotype terms form a directed acyclic graph (DAG) and are connected by directed "IS\_A" edges (denoting subclass relationships). 
The files used in this study are "hp.obo" and "phenotype.hpoa", with the version being hp/releases/2023-06-06.

Orphanet \footnote{https://www.orpha.net/consor/cgi-bin/index.php}, funded by the French Ministry of Health, is a non-profit online resource and information platform focusing on rare diseases. It offers information on over 6,000 rare diseases, written by medical experts and researchers and subjected to rigorous quality review. This information includes etiology, symptoms, diagnostic methods, treatment options, and prognosis. Our work primarily utilizes the annotation information on rare diseases and genes from this platform.

Online Mendelian Inheritance in Man (OMIM) \footnote{https://www.omim.org/} is a comprehensive database of genes and genetic diseases, collecting and organizing extensive information about human genetic disorders. It includes descriptions of over 20,000 genetic diseases, covering genetic etiology, clinical manifestation, inheritance patterns, molecular mechanisms, and related literature.

The National Rare Disease Registry System of China (NRDRS) \footnote{https://www.nrdrs.org.cn/xhrareweb/homeIndex}, overseen by Peking Union Medical College Hospital, is a national-level online registry platform for rare diseases, aimed at establishing unified rare disease registration standards and norms. The first list of the Compendium of China’s Rare Disease (CCRD) compiles detailed information on 144 rare diseases. The annotations of relationships between diseases and phenotypes are manually extracted. The version used in our study is 2019-11.

\subsubsection{Comparison Analysis of Node2vec and IC value-based random walk}

In our integrated rare disease knowledge graph, we implement an IC value-based random walk algorithm to obtain embeddings for phenotype and disease nodes within the graph. While Node2Vec utilizes two parameters, \( p \) and \( q \), to control the walking probability, we dynamically adjust this based on the IC values of different phenotypes. The settings for other parameters are as follows:

\begin{framed}
\footnotesize
\noindent'embedding\_dim': 256,
    
\noindent'walk\_length': 45,

\noindent'context\_size': 35,

\noindent'walks\_per\_node': 40,

\noindent'num\_negative\_samples': 1,

\noindent'sparse': True,

\noindent'loader\_batch\_size': 256,

\noindent'loader\_shuffle': True,

\noindent'loader\_num\_workers': 4,

\noindent'learning\_rate': 0.01,

\noindent'epoch\_nums': 36

\end{framed}

\end{document}